\documentclass[twoside,11pt]{article}

\usepackage[preprint]{jmlr2e}

\usepackage{amsmath,amsfonts,amssymb}
\usepackage{dsfont}

\newcommand\independent{\protect\mathpalette{\protect\independenT}{\perp}}
\def\independenT#1#2{\mathrel{\rlap{$#1#2$}\mkern2mu{#1#2}}}

\newcommand{\R}{\mathbb{R}}
\newcommand{\N}{\mathbb{N}}

\newcommand{\E}{\mathbb{E}}

\newcommand{\Nor}{\mathcal{N}}
\DeclareMathAlphabet{\pazocal}{OMS}{zplm}{m}{n}
\newcommand{\unif}{\pazocal{U}}
\newcommand{\bSigma}{\boldsymbol{\Sigma}}
\newcommand{\bhSigma}{\boldsymbol{\hat{\Sigma}}}

\newcommand{\bmu}{\boldsymbol{\mu}}

\newcommand{\btheta}{\boldsymbol{\theta}}

\newcommand{\bphi}{\boldsymbol{\phi}}

\newcommand{\bO}{\boldsymbol{0}}
\newcommand{\bI}{\boldsymbol{I}}
\newcommand{\ELBO}{\mathcal{L}}

\newcommand{\KL}{D_{KL}}

\newcommand{\Ind}{\mathds{1}}

\newcommand{\bhmu}{\hat{\bmu}}

\newcommand{\diag}{\textnormal{diag}}

\usepackage{todonotes}
\usepackage{booktabs}
\usepackage{hyperref}
\usepackage{blindtext}

\title{Estimating the Value-at-Risk by Temporal VAE}

\author{%
	\name Robert Sicks\;\thanks{Corresponding author} \email robert.sicks@itwm.fraunhofer.de\\
	\addr Department of Financial Mathematics\\
	Fraunhofer ITWM\\
	Kaiserslautern
	\AND
	\name Stefanie Grimm \\
	\addr Department of Financial Mathematics\\
	Fraunhofer ITWM\\
	Kaiserslautern
	\AND
	\name Ralf Korn \\
	\addr Department of Financial Mathematics\\
	TU Kaiserslautern\\
	Kaiserslautern
	\AND
	\name Ivo Richert \\
	\addr Department of Financial Mathematics\\
	Fraunhofer ITWM\\
	Kaiserslautern
}

\begin{document}

	\maketitle
	
	\begin{abstract}%
		Estimation of the value-at-risk (VaR) of a large portfolio of assets is an important task for financial institutions. As the joint log-returns of asset prices can often be projected to a latent space of a much smaller dimension, the use of a variational autoencoder (VAE) for estimating the VaR is a natural suggestion. To ensure the bottleneck structure of autoencoders when learning sequential data, we use a temporal VAE (TempVAE) that avoids an auto-regressive structure for the observation variables. However, the low signal-to-noise ratio of financial data in combination with the auto-pruning property of a VAE typically makes the use of a VAE prone to posterior collapse. Therefore, we propose to use annealing of the regularization to mitigate this effect. As a result, the auto-pruning of the TempVAE works properly which also results in excellent estimation results for the VaR that beats classical GARCH-type and historical simulation approaches when applied to real data. 
	\end{abstract}
	\vspace{0.3cm}
	
	\begin{keywords}
		Value-at-Risk, variational autoencoders, recurrent neural networks, risk-management, auto-pruning, posterior collapse
	\end{keywords}
	
	\section{Introduction}
	

	The value-at-risk (VaR) is the most prominent risk measure in finance. Being defined as a typically high quantile of the loss function of a portfolio of assets over a fixed time horizon, it is notoriously hard to estimate. Methods to determine it range from simple historical simulation to parametric approximation, in particular variants of  GARCH-models combined with conditional normal distribution assumptions (see e.g. \cite{Engle2012}), to first applications of neural networks (see e.g. \cite{Chen2009}).
	
	In this work, we propose the temporal variational autoencoder (TempVAE), a variational recurrent neural network (VRNN) designed to handle the low signal-to-noise ratio in financial time series, and apply it to VaR estimation. By leaving out an auto-regressive structure for the observation variables, we explicitly force the model to use the bottleneck structure from autoencoders. 
	
	The key concept behind the TempVAE is the variational autoencoder (VAE; see \cite{Kingma2014} and \cite{Rezende2014}). VAE produce stochastic latent representations as opposed to standard autoencoders (AE) which yield deterministic ones. The result of the distribution model is a more meaningful latent space fragmentation and an increase in robustness against common training pitfalls such as over-fitting or model misspecification. 
	
	With the introduction of VRNNs, \cite{Chung2015} show promising results to model complex data sequences, like speech data, as compared to standard RNN models. Based on this work, \cite{Luo2018a} propose a neural stochastic volatility model (NSVM) and show that, using VRNN, it is possible to adequately model the volatility of financial data. However, the low signal-to-noise ratio makes variational models prone to posterior collapse, i.e. the encoder becoming independent of the input.
	
	The reason for the posterior collapse is the auto-pruning. During training the net sets nodes for the latent space
	inactive. This property of variational models is known as a double-edged sword and has been analyzed by various authors (see e.g. \cite{Burda2016} and \cite{Sicks2021}). On the one hand, the model focuses on useful representations by pruning away not needed nodes. On the other hand, it is considered a problem when too many nodes become inactive before learning a useful representation. 
	
	As a variational model, the TempVAE is also prone to posterior collapse. We propose to use annealing (see also \cite{Bowman2015}) of the regularization to mitigate this effect. This way, we benefit from the auto-pruning property to find useful representations of sequence data.

	Using our new proposed model, we make the following contributions:
	\begin{enumerate}
		\item We show that the TempVAE possesses the auto-pruning property. Given synthetic data, the TempVAE identifies the correct number of latent factors to adequately model the data.
		\item For the VaR estimation, we show that our model performs excellent and beats the benchmark models.
	\end{enumerate}

	The remaining paper is structured as follows. In Section \ref{sec:Comparison to Related Work}, we present related work and compare it to our research. The definition of the TempVAE can be found in Section \ref{sec:The Temporal Variational Autoencoder}. In Section \ref{sec:Implementation and Experiments}, we explain the data and provide our results for the VaR estimation with the TempVAE. We conclude the paper in Section \ref{sec: Conclusion}.
	
	\section{Comparison to Related Work}\label{sec:Comparison to Related Work}
	
	Various authors have considered stochastic units in RNNs. \cite{Bayer2014}, \cite{Chung2015} and \cite{Fraccaro2016} propose stochastic RNN with auto-regressive structures for the observations while \cite{Goyal2017} and \cite{Luo2018a} extend the approach of \cite{Chung2015} via auxiliary cost terms in the objective or bidirectional RNNs respectively. Moreover \cite{Xu2021} propose a similar model to \cite{Luo2018a} and apply it to financial return data.

	\cite{Krishnan2017} consider a Gaussian state space model by assuming the Markov property for the latent variables and excluding auto-regressive structures. Given their assumptions, they show that the true posterior for the latents given the observations only depends on future observations. Additionally, \cite{Fraccaro2017} consider a combination of VAE with a state space model. Specifically, they assume that the observations of the state space model make up the bottleneck of the VAE. This way, they achieve a disentanglement of the latent representation and the dynamics.
	
	Finally, \cite{Bowman2015} consider the encoding and decoding of single layer RNN sequences via VAEs to model speech data. They argue that annealing of the regularization term is needed in order to learn meaningful information passing through the bottleneck.
	
	Alternatively to the models by \cite{Chung2015} and \cite{Luo2018a}, we model a temporal dependency only in the prior distribution. Therefore, our model is similar to the deep Markov model by \cite{Krishnan2017}, but we do not use the Markov assumption. 
	
	By excluding the auto-regressive dependency for the observation variables, the return distribution is independent from past returns, but not from past latents. Therefore, we force the encoder of the implemented model to encode as much information as possible in the latents $Z$. As a consequence, properties of conventional VAE like the auto-pruning can work properly. This is not necessarily the case if we model the decoder distribution to be auto-regressive\footnote{e.g.
		$p(R|Z)= \prod_{t=1}^{T} p(R_t | R_{1:t-1},Z_{1:t})$} on the returns $R$, 
	as assumed by \cite{Luo2018a} and \cite{Chung2015}. In this case, a substantial part of the information that can be used by the decoder wouldn't have to pass the bottleneck. Even if the posterior $q(Z|R)$ collapses and becomes independent of the input, the decoder would be able to use past observed $R$ to achieve a good fit.
	


	Various authors have proposed to estimate the VaR with artificial neural networks (ANN).
	\cite{Liu2005} uses the estimates of historical simulation and GARCH as input for ANNs. Therefore his model can be interpreted as an ensemble model. \cite{Chen2009} and \cite{Arimond2020} estimate the VaR by modelling parameters of their respective distribution assumption. \cite{Chen2009} use a standard ANN whereas \cite{Arimond2020} compare ANNs with temporal convolutional neural nets and RNNs. \cite{Arian2020} use standard VAE for the task of estimating the VaR. To account for the time dependency in the data, they preprocess the data with rolling window statistics. 
	
	By using the TempVAE for this task, we have a time-dependent and, due to the auto-pruning property, parsimonious model for the data sequences. As the joint log-returns often can be projected to a lower dimensional space, estimation of the VaR with this model comes as a natural task.


	\section{The Temporal Variational Autoencoder}\label{sec:The Temporal Variational Autoencoder}

\begin{figure}
		\center
		\includegraphics[scale=0.7]{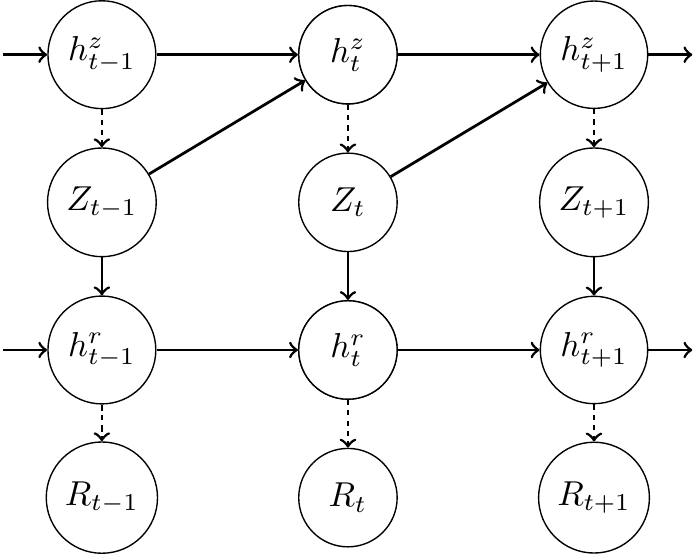}
		\caption{The dependency structure of the generative model. Realizations of $Z$ influence the realizations of $R$ as well as all future realizations of $R$ and $Z$. The Gaussian distribution parameters from the MLP as well as the sampling are represented by the dashed lines.
			Further, information of former time points is propagated through the RNN hidden states $h^r_{\cdot}$ and $h^z_{\cdot}$.}
		\label{fig:tikzit_image_TempVAE}
		
	\end{figure}

	We use the TempVAE to learn the multivariate distribution of asset returns. Therefore, we assume that the return series $R=R_{1:T}$ is generated via latent variables $Z=Z_{1:T}$, where $T\in\N$ and $T>1$. The distributions of the variables $Z_t$ are estimated via approximate inference and we expect these to hold information like the current market environment, peer group behaviour or interactions. Given the two discrete-time stochastic processes $R$ and $Z$, we assume an auto-regressive time dependency on the latents, given by
	\begin{align}
	Z_t|Z_{1:t-1} &\sim \Nor\left(\bmu^z(Z_{1:t-1}),\bSigma^z(Z_{1:t-1})\right),\label{eq:assumption z}\\
	R_t|Z_{1:t} &\sim \Nor\left(\bmu^r(Z_{1:t}),\bSigma^r(Z_{1:t})\right),\label{eq:assumption r}
	\end{align}
	where $\bmu^z$, $\bmu^r$, $\bSigma^z$ and $\bSigma^r$ are functions mapping to $\R^\kappa$, $\R^d$, $\R^{\kappa \times \kappa}$ and $\R^{d\times d}$ respectively. Given this dependency structure, the joint distribution can be factorized as
	\begin{align}
	p_{\btheta}(Z) &= \prod_{t=1}^{T} p_{\btheta}(Z_t | Z_{1:t-1}), \label{eq:joint fact z}\\
	p_{\btheta}(R|Z) &= \prod_{t=1}^{T} p_{\btheta}(R_t | Z_{1:t})\label{eq:joint fact r|z}.
	\end{align}
	To account for the dependency of the past sequences, we use RNN layers. Further do we use Multi-Layer Perceptrons (MLPs) to map from the output of the RNNs to the parameter space of the distributions. In total, we use the model architecture given by
	\begin{align}
	\left\{\bmu_t^z,\bSigma_t^z\right\} &= \mathrm{MLP}^z_G\left(h_t^z\right), \label{eq:prior RNN MLP 1}\\
	h_t^z &= \mathrm{RNN}_G^z\left(h_{t-1}^z,Z_{t-1}\right), \label{eq:prior RNN MLP 2}\\
	Z_{t} &\sim \Nor\left(\bmu_t^z,\bSigma_t^z\right),\label{eq:prior RNN MLP 3}\\
	\nonumber\\
	\left\{\bmu_t^r,\bSigma_t^r\right\} &= \mathrm{MLP}^r_G\left(h_t^r\right),\label{eq:generative RNN MLP 1}\\
	h_t^r &= \mathrm{RNN}_G^r\left(h_{t-1}^r,Z_{t}\right),\label{eq:generative RNN MLP 2}\\
	R_{t} &\sim \Nor\left(\bmu_t^r,\bSigma_t^r\right)\label{eq:generative RNN MLP 3},
	\end{align}
	which is also depicted in Figure \ref{fig:tikzit_image_TempVAE}. Combining the hidden states of the model is possible and yields a more general model formulation. But during our experiments, we found that it is favourable to have the prior as not trainable (see Appendix \ref{app:Analysing trainable prior}). In order to achieve this, we separate the hidden states as shown in Figure \ref{fig:tikzit_image_TempVAE}.
	
	\noindent We approximate the intractable $p_{\btheta}(Z|R)$ by the inference distribution $q_{\bphi}(Z|R)$. 
	Similar to \cite{Krishnan2017}, one can show that $Z_t \independent R_{1:t-1} | Z_{1:t-1}$. Therefore, we can factorize the distribution such that the variable $Z_t$ depends on the (future) observations $R_{t:T}$:
	\begin{equation}\label{eq:p posterior z|r future dependency}
	p_{\btheta}(Z|R) = \prod_{t=1}^{T} p_{\btheta}(Z_t | Z_{1:t-1}, R_{t:T}).
	\end{equation}
	\cite{Krishnan2017} argue that $q$ should be factorized the same way. As $q$ is only an approximation to $p$, we also considered providing the full sequence which yields a better performance (see Appendix \ref{app:Encoder depencency}). We therefore factorize $q$ as
	\begin{equation}\label{eq:inference z|r}
	q_{\bphi}(Z|R) = \prod_{t=1}^{T} q_{\bphi}(Z_t | Z_{1:t-1},R_{1:T})
	\end{equation}
	and assume $q_{\bphi}(Z_t | Z_{1:t-1},R_{1:T}) \sim \Nor\left(\bhmu^z(Z_{1:t-1},R_{1:T}),\bhSigma^z(Z_{1:t-1},R_{1:T})\right)$. $\bhmu^z$ and $\bhSigma^z$ are functions mapping to $\R^\kappa$ and $\R^{\kappa \times \kappa}$ respectively. Furthermore, we implement the inference distribution via a bidirectional RNN as follows
	\begin{align}
	\left\{\bhmu_t^z,\bhSigma_t^z\right\} &= \mathrm{MLP}^z_I\left(\hat{h}_t^z\right)\label{eq:encoder RNN MLP 1}\\
	\hat{h}_t^z &= \mathrm{RNN}_I^z\left(\hat{h}_{t-1}^z,Z_{t-1},\left[\hat{h}_t^{\rightarrow},\hat{h}_t^{\leftarrow}\right]\right) \label{eq:encoder RNN MLP 2}\\
	\hat{h}_t^{\rightarrow} &= \mathrm{RNN}_I^z\left(\hat{h}^{\rightarrow}_{t-1},R_{t-1}\right)\label{eq:encoder RNN MLP 3}\\
	\hat{h}_t^{\leftarrow} &= \mathrm{RNN}_I^z\left(\hat{h}^{\leftarrow}_{t+1},R_{t+1}\right)\label{eq:encoder RNN MLP 4}\\
	Z_{t} &\sim \Nor\left(\bhmu_t^z,\bhSigma_t^z\right).\label{eq:encoder RNN MLP 5}
	\end{align}
	
	\noindent As we have a distribution assumption with an intractable $p(Z|X)$, the usual suspect for the model training is the evidence lower bound (ELBO) given as
	\begin{align}
	\ELBO(\btheta,\bphi) &= \E_{q_{\bphi}(Z|R)}\left[\log p_{\btheta}(R|Z)\right] +  \E_{q_{\bphi}(Z|R)}\left[\log \left(\dfrac{p_{\btheta}(Z)}{q_{\bphi}(Z|R)}\right)\right]\\
	&= \E_{q_{\bphi}(Z|R)}\left[\log p_{\btheta}(R|Z)\right] - \KL\left(q_{\bphi}(Z|R) || p_{\btheta}(Z) \right).
	\end{align}
	The components are given by equations \eqref{eq:joint fact z}, \eqref{eq:joint fact r|z} and \eqref{eq:inference z|r}.
	As we have $\log p_{\btheta}(R) \geq \ELBO(\btheta,\bphi)$, maximization of the ELBO most likely leads to an increase in the likelihood of our distribution model. In a similar way to \cite{Krishnan2017}, we can rewrite the ELBO as
	\begin{align}\small \ELBO(\btheta,\bphi) = \E_{q_{\bphi}(Z|R)}\left[\log p_{\btheta}(R|Z)\right] - \sum_{t=1}^{T} \E_{q_{\bphi}(Z_{1:t-1}| R)} \Big[\KL\big(q_{\bphi}(Z_t| Z_{1:t-1},R) \: \big\lVert \: p_{\btheta}(Z_t| Z_{1:t-1}) \big)\Big]. \label{eq: Xu KL ELBO}
	\end{align}
	Since we assume a Gaussian distribution for $q$ and $p$, we can analytically calculate the Kullback-Leibler divergence (KL-Divergence) inside the expectation in \eqref{eq: Xu KL ELBO}. We approximate the outer expectations numerically by Monte Carlo.
	

	\subsection{The Auto-Pruning Property of VAEs and the Posterior Collapse}\label{sec: beta annealing and auto pruning}
	
	The auto-pruning of VAE originates from the KL-Divergence and has been analyzed by various authors (see \cite{Burda2016} and \cite{Sicks2021}). When nodes in the bottleneck are ``pruned away'', it means that these nodes are not used in the neural net and therefore also not for the reconstruction. The auto-pruning capabilities of VAE are desirable, as the model focuses on useful latent representations. However, it is considered a problem when too many units become inactive before learning such a useful representation. In the extreme case, the KL-Divergence between $q_{\bphi}(Z| R)$ and $p_{\btheta}(Z)$ is 0 and we have $q_{\bphi}(Z| R)\equiv p_{\btheta}(Z)$. $q$ is independent of $R$ and we say that the posterior $q$ collapses as the input does not matter.
	
	As we consider multivariate Gaussian distributions with diagonal covariance, it becomes apparent that only a subset of nodes can be inactive. The posterior does not collapse in this case, as information can still be propagated to the decoder through active nodes.
	
	When using the TempVAE on financial data, we observe an instantaneous collapse of the posterior at the beginning of the training. In our point of view, the signal within the financial returns data is not strong enough to be captured before the KL-Divergence term dominates the ELBO and causes a posterior collapse. Therefore, we consider $\beta$-annealing of the KL-Divergence similar to \cite{Bowman2015} (see Appendix \ref{app: model implementation and beta annealing}).
	
	To measure the amount of active units, \cite{Burda2016} propose an activity statistic that is calculated after training. For each node, they estimate the activity by 
	\[
	A_{u} = \mathrm{Cov}_x\left(\E_{q_{\phi}\left(u|x\right)}[u]\right)
	\]
	and call a node inactive, if $A_{u} < 0.01$. Since we assume a time-dependent model for the prior in \eqref{eq:joint fact z}, we adjust the activity statistic to calculate the activities of a sequence of $Z_{1:M} \in \R^{M \times \kappa}$. For $m=1,\ldots,M$ and $k=1,\ldots,\kappa$ we calculate
	\begin{equation}\label{eq: time-dep activity}
	A_{z}^{(m,k)} := \mathrm{Cov}_{R,Z_{1:m-1}} \left(\E_{q_{\phi}\left(Z_{m,k}|Z_{1:m-1},R\right)}\left[Z_{m,k}\right] - \E_{p_{\theta}\left(Z_{m,k}|Z_{1:m-1}\right)}\left[Z_{m,k}\right]\right).
	\end{equation}
	The statistic becomes equal to the one proposed by \cite{Burda2016}, if we assume a time-independent standard Gaussian prior\footnote{Then, the assumption is $p_{\btheta}(Z) = \prod_{t=1}^{T} p_{\btheta}(Z_t)$, with $p_{\btheta}(Z_t) \sim \Nor(\bO, \bI)$ for all $t=1,\ldots,T$. } in \eqref{eq:joint fact z}. With the statistic in \eqref{eq: time-dep activity} we quantify the effect of the input $R$. Similar to \cite{Burda2016}, we call a node inactive from the input, if $A_{z}^{(m,k)} < 0.01$. 

	\section{Implementation and Experiments}\label{sec:Implementation and Experiments}
	
	In this section, we present the results we achieved on four data sets (see Sections \ref{sec:data and preprocessing}) with our model regarding the identification of the underlying signal (see Section \ref{sec:experiments activity}). Then, in Section \ref{sec: model fit and risk management} we present results for the fit to financial data as well as the VaR estimation. For further details on the model implementation and for an additional ablation study validating our design choices see Appendices \ref{app: model implementation and beta annealing} and \ref{app:Further ablation studies}.

	\subsection{Description of employed Data Sets}\label{sec:data and preprocessing}
	
	We consider the four data sets ``DAX'', ``S\&P500'', ``Noise'' and ``Oscillating PCA'' for our studies. Throughout the section, data is stored in $d$-dimensional vectors
	\begin{equation}\label{eq: d-dimensional data series}
	R_{t} := \left(R_{t,1} ,\ldots,R_{t,d} \right)^T \in \R^{d}.
	\end{equation}
	First, we consider ($d$=22) stock price time series, provided by DAX (German stock index) enlisted companies, from 11.06.2001 to 09.06.2021. We calculate the daily logarithmic returns\footnote{We consider log-returns as these are better suited with a Gaussian assumption. This is common practice when modeling financial data. For our later evaluation on the risk management, we therefore transform the log-return forecasts of the models to actual returns (by using $R_{t,i}^\text{nonlog}:= \exp\left(R_{t,i}\right) - 1$) to model the extremes in the data adequately. } via the observed close prices. Therefore, if $S_{t,i}$ denotes the price of stock $i=1,\ldots,d$ at time $t=1,\ldots,T$, the data set ``DAX'' contains
	\begin{equation}\label{eq: log returns}
	R_{t,i} := \ln\left(S_{t,i}\right) - \ln\left(S_{t-1,i}\right).
	\end{equation}
	For ``S\&P500'', we consider 397 stocks of the S\&P500 index that have a history from 02.01.2002 up to 09.06.2021. We calculate the log-returns analogously to the ``DAX'' data.
	
	The data set ``Noise'' is just white noise with dimension $d=22$. Hence, for all $t=1,\ldots,T$ and $i=1,\ldots,d$
	\[
	R_{t,i} \sim \Nor\left(0,1\right).
	\]
	
	\begin{figure}
		\center
		\includegraphics[scale=0.75]{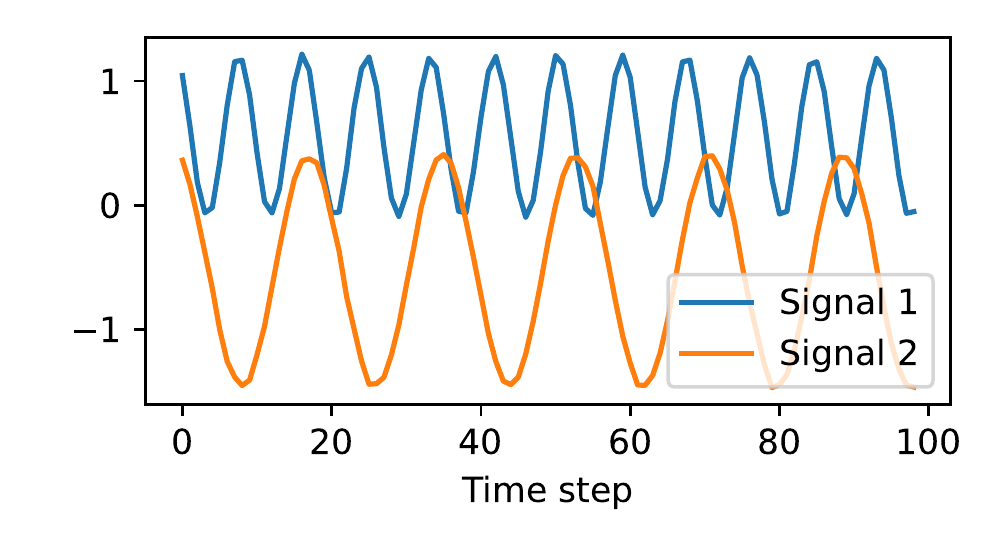}
		\caption{The first 100 steps of two oscillating signals. The two signals have different amplitude, frequency and origin. }
		\label{fig:osc_pca_signals}
		
	\end{figure}
	
	\noindent The data set ``Oscillating PCA $k$'' (with $k \in \{2,5,10\}$) is constructed based on $k$ independent harmonic oscillator signals. Each univariate signal $Z_{1:T}$ is constructed by randomly choosing an intercept $i\sim\unif\left(-1,1\right)$, an amplitude $a\sim\unif\left(-1,1\right)$, a frequency $f\sim\unif\left(0.5,24\right)$ and a noise series $\epsilon_{1:T}$, with $\epsilon\sim \Nor\left(0, 0.02^2\right)$. Then, we set 
	\[
	Z_t = i + a \cdot \cos\left(\dfrac{t}{100} \cdot f \cdot \pi \right) + a \cdot \epsilon_t.
	\]
	The case $k=2$ is shown in Figure \ref{fig:osc_pca_signals}. The oscillator sequences $Z_{1:T,1:k} \in \R^{T\times k}$ are then transformed to a 22-dimensional series by applying a randomly chosen rotation $U \in \R^{d \times k}$:
	\[
	S_t = U \cdot Z_{t,1:k}^T +5 \qquad \text{and} \qquad 	R_{t,i} := \ln\left(S_{t,i}\right) - \ln\left(S_{t-1,i}\right)
	\]
	for each $t=1,\ldots,T$ and $i=1,\ldots,d$. 
	
	Finally, a sliding window is applied to the $d$-dimensional series $R_{1:T}$ to generate $T-M$ consecutive series in $\R^{d\times M}$ of size $M = 21$. For further details on preprocessing procedures, see Appendix \ref{app: model implementation and beta annealing}.
	
	\subsection{Signal Identification}\label{sec:experiments activity}
	
\begin{figure}
		\center
		\includegraphics[scale=0.55]{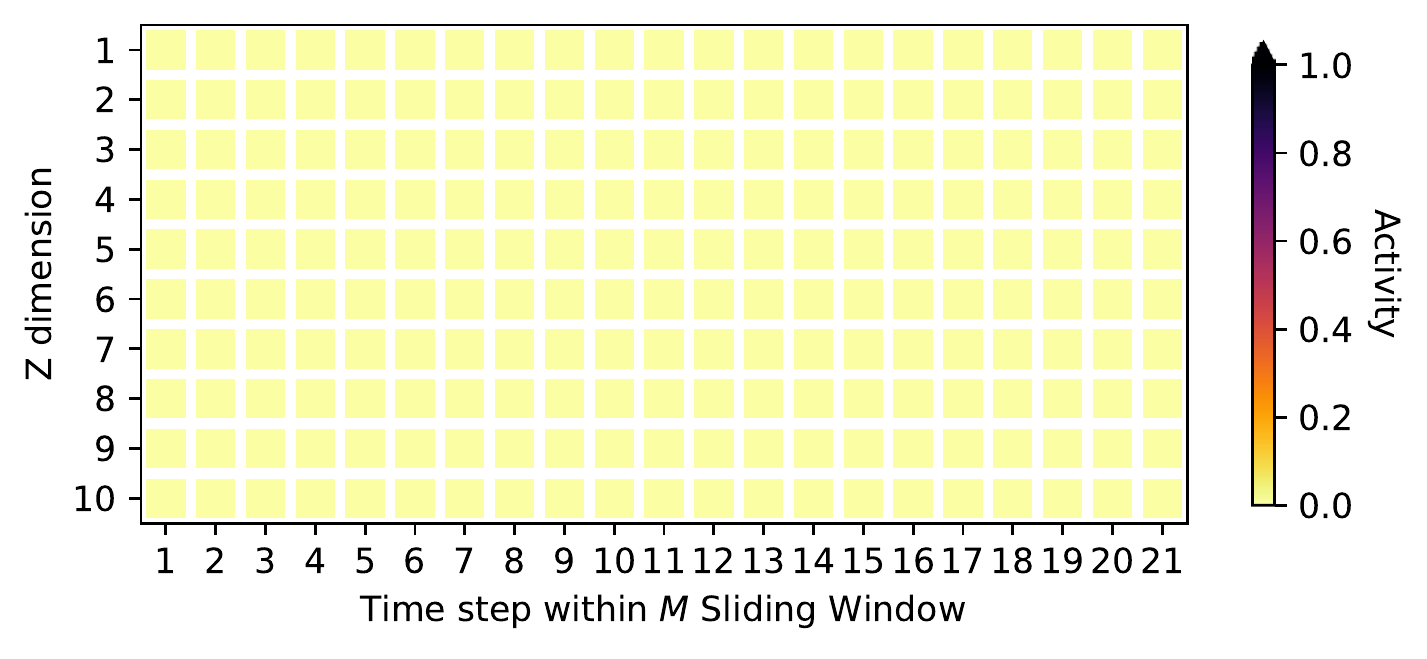}
		\caption{The activity statistics for the ``Noise'' data. We consider sequences of size $M=21$ as input. Depicted are the activity values of statistic \eqref{eq: time-dep activity} for the $\kappa=10$ dimensional latent space. }
		\label{fig:plot_Activities_heatmap_noise}
		
	\end{figure}

\begin{figure}
		\center
		\includegraphics[scale=0.55]{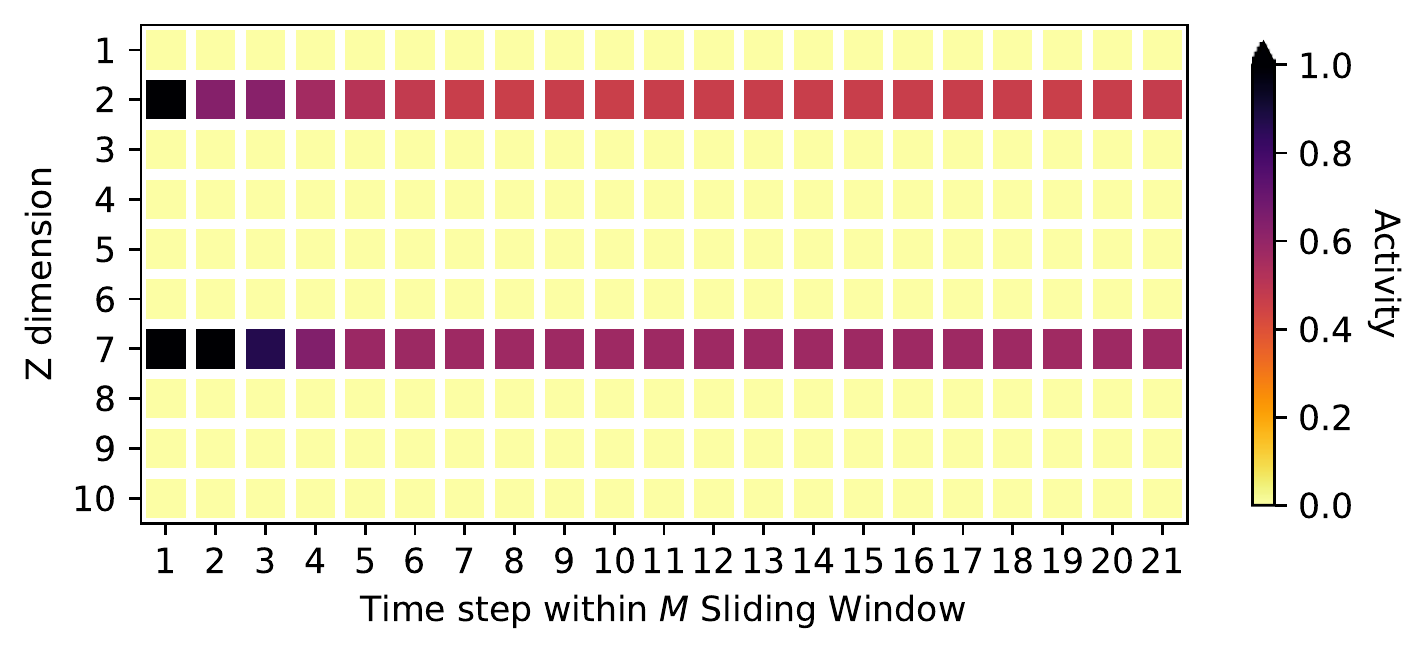}
		\caption{The activity statistics for the ``Oscillating PCA 2'' data. We consider sequences of size $M=21$ as input. Depicted are the activity values of statistic \eqref{eq: time-dep activity} for the $\kappa=10$ dimensional latent space. }
		\label{fig:plot_Activities_heatmap_osc_pca}
		
	\end{figure}

	We train the TempVAE on the four data sets from Section \ref{sec:data and preprocessing} and analyze calculated activities given by \eqref{eq: time-dep activity} as well as the fit to the data. As we can see in Figure \ref{fig:plot_Activities_heatmap_noise} and \ref{fig:plot_Activities_heatmap_osc_pca}, the TempVAE correctly identifies the amount of hidden dimensions. For the ``Noise'' data, no activity is reported at all. The decoder model $p(R|Z)$ is left on it's own to model the data as no meaningful information can be identified that can be propagated from input to output. For the data ``Oscillating PCA 2'', the model correctly identifies the two active nodes per time step. 	
	For the ``Oscillating PCA 5'' and ``Oscillating PCA 10'' data, the model also identifies two active dimensions through time (see Appendix \ref{app:TempVAE latent space act osc pca 5 10}). Though this is not the amount given by the linear construction, identifying less dimensions is not necessarily problematic since we have a non-linear model. We observe comparable fitting results for these data sets (see Appendix \ref{app:TempVAE latent space act osc pca 5 10}).
	
	Applied to the financial data, we can observe two and four active dimensions for the ``DAX'' and ``S\&P500'' data sets respectively (see Appendix \ref{app: Activities Stock Market data and High dim model}). The fact that only a few active nodes could be identified is not surprising. In their work \cite{Laloux2011} show that only a small fraction of dimensions (roughly 11) are needed to linearly model 406 assets of the S\&P500 during the years 1991-1996.
	
	

	\subsection{Fit to Financial Data and Application to Risk Management}\label{sec: model fit and risk management}
	
	To assess the performance of the model on the ``DAX'' data\footnote{The benchmark models are note feasible on the ``S\&P500'' data due to a too high dimension. }, we compare the TempVAE to a GARCH model as well as to the multivariate GARCH versions DCC-GARCH-MVN and DCC-GARCH-MVt (see Appendix \ref{app:GARCH and MGARCH} for details). Therefore, we calculate the negative log-likelihood (NLL), given by
	\begin{equation*}
	\mathrm{NLL}(R_t^\text{nonlog},\mu_t,\Sigma_t) := \frac{1}{2}\left(\kappa \log\left(2 \pi\right) + \log\left|\Sigma_t\right| + \left(R_t^\text{nonlog} - \mu_t \right)^T \Sigma_t^{-1} \left(R_t^\text{nonlog} - \mu_t \right) \right)
	\end{equation*}
	and average across all timestamps in the test set.	$R_t^\text{nonlog}$ are the non-logarithmic returns of the test data and $\mu_t$ as well as $\Sigma_t$ are estimated by the models. To estimate these values, we first sample the log-returns from a model. For the TempVAE, we estimate these empirically by sampling
	\begin{align*}
	z_{1:t}^{(i)} \sim q_{\phi}(Z|r_{1:t-1}) \quad \text{ as well as } \quad r_t^{(i)} \sim p_{\btheta}(R|z_{1:t}^{(i)} ) &
	\end{align*}
	for $i=1,\ldots,1000$. Then, we transform these samples to nonlog-returns and calculate the empirical means and covariances to approximate $\mu_t$ and $\Sigma_t$. 
	
	We also consider the diagonal NLL, where we use the diagonal matrix with entries $\diag\left(\Sigma_t\right)$ instead of $\Sigma_t$. Further do we calculate a univariate version by considering the returns of an equally weighted portfolio containing the 22 assets. Given the log-returns in \eqref{eq: log returns}, we calculate these by
	\begin{equation}\label{eq: portfolio returns}
	R_{t,i}^{P,\text{nonlog}} := \dfrac{1}{d} \sum_{i=1}^{d} \left(\exp\left(R_{t,i}\right) - 1\right).
	\end{equation}
	
	\begin{table}[h!]
		\center
		\begin{tabular}{l|rrr}
			Model & Diagonal NLL & NLL & Portfolio NLL \\ \hline
			GARCH & \textbf{22.76} & 22.84 & -0.51 \\ 
			TempVAE & 25.79 & 20.35 & -3.04 \\ 
			DCC-GARCH-MVN & 25.03 & \textbf{18.69} & \textbf{-3.07} \\ 
			DCC-GARCH-MVt & 25.44 & 19.17 & -3.05 \\ 
		\end{tabular}
		\caption{Fit scores for models TempVAE, GARCH and two multivariate GARCH. In all cases, lowest is best. }
		\label{tab: NLL comparison}
	\end{table}
	
	Table \ref{tab: NLL comparison} displays the different fit scores for the three models. 
	Considering the NLL scores, the GARCH and normal DCC-GARCH models achieve the best performance since these models are directly minimizing the respective scores. The TempVAE as a regularized model has to find a trade-off between the NLL and the regularization terms. Nonetheless, if we consider the scores for the portfolio returns, we see all multivariate models performing similarly and outperforming the GARCH model. Due to the strong dependency between the different assets in the portfolio, estimating the underlying covariance structure plays an essential role in estimating the next day's portfolio performance. Further, we could observe the models capturing the correlations between the assets (see Appendix \ref{app: VaR plots} for an excerpt of the first two dimensions). 
	
	To compare the performance on the VaR estimation, as a further method we consider historical simulation (HS).\footnote{We consider a time window of 180 days. } We calculate the VaR estimates via 1000 Monte Carlo samples from the respective distributions.\footnote{Note that usually one can calculate the VaR estimates for GARCH models analytically. But as we apply a non-linear transform of the modeled log-returns (see expression \eqref{eq: portfolio returns}) this task is not trivially performed.} The VaR95 is then the 50th of the ordered samples and the VaR99 the 10th sample. We evaluate these with the `Regulatory Loss Function' (RLF) proposed by \cite{Sarma2003} which penalizes exceedance of the VaR and is given by
	\[
	\mathrm{RLF}\left(\textnormal{VaR}_t,r_t\right)= 
	\begin{cases}
	\left(\textnormal{VaR}_t - r_t\right)^2,& \text{if } r_t\geq \textnormal{VaR}_t\\
	0,              & \text{otherwise.}
	\end{cases}
	\]
	We calculate the average of these values over the test set.
	
	\begin{table}[h!]
		\center
		\begin{tabular}{l|rrrrrr}
			Model & RLF95 & RLF99 & Br95 & Br99 \\ \hline
			GARCH & 44.47 & 32.35 & 24.3 & 17.8 \\ 
			TempVAE & 12.64 & 5.96 & 4.8 & \textbf{1.3} \\ 
			DCC-GARCH-MVN & 13.15 & 7.10 & 5.4 & 2.1 \\
			DCC-GARCH-MVt & 10.23 & 4.14 & 3.9 & 0.6 \\ 
			HS & 14.57 & 7.43 & \textbf{5.1} & \textbf{1.3} \\ 
		\end{tabular}
		\caption{The table displays the values (all of these have to be multiplied by $10^{-6}$) of the RLF scores as well as the average count of breaches for the VaR95 and the VaR99 within the test set. For the RLF values, lowest is best. For the average breaches, the respective quantile assumption is the optimal value. Hence for VaR95 the best value is 5 and for VaR99 the best value is 1. 
		}
		\label{tab: VaR scores}
	\end{table}
	
	\noindent In Table \ref{tab: VaR scores}, the losses for the models as well as the average number of breaches are displayed. The amount of average breaches helps us to validate the VaR estimates by their 5\% and 1\% quantile assumption and a good performance is a necessary condition. As the RLF can be simply minimized to zero by decreasing any VaR estimates, we only compare models in their RLF that show a reasonable amount of breaches. Therefore, though DCC-GARCH-MVt has the lowest values for the RLF95 and RLF99, the TempVAE produces the best estimates. 
	
	In Figure \ref{fig:VaR95 MGARCH,NSVM inset}, the VaR95 values for the DCC-GARCH-MVN, HS and TempVAE are displayed. The full path of the VaR95 estimates as well as the VaR99 estimates can be found in Appendix \ref{app: VaR plots}. We can see the VaR95 estimates for the DCC-GARCH-MVN and TempVAE are quickly responding to the changes in the volatility of the data. Comparing TempVAE directly to DCC-GARCH-MVN, we see the VaR estimates to be more volatile but also responsive to the evolution of the returns. The HS on the other side is showing the well-known time-delay (see \cite{Liu2005}). Even though in terms of average breaches the HS is best followed by TempVAE, in our point of view the TempVAE provides the most reasonable estimates among the models.  
	\begin{figure}
		\center
		\includegraphics[scale=0.8]{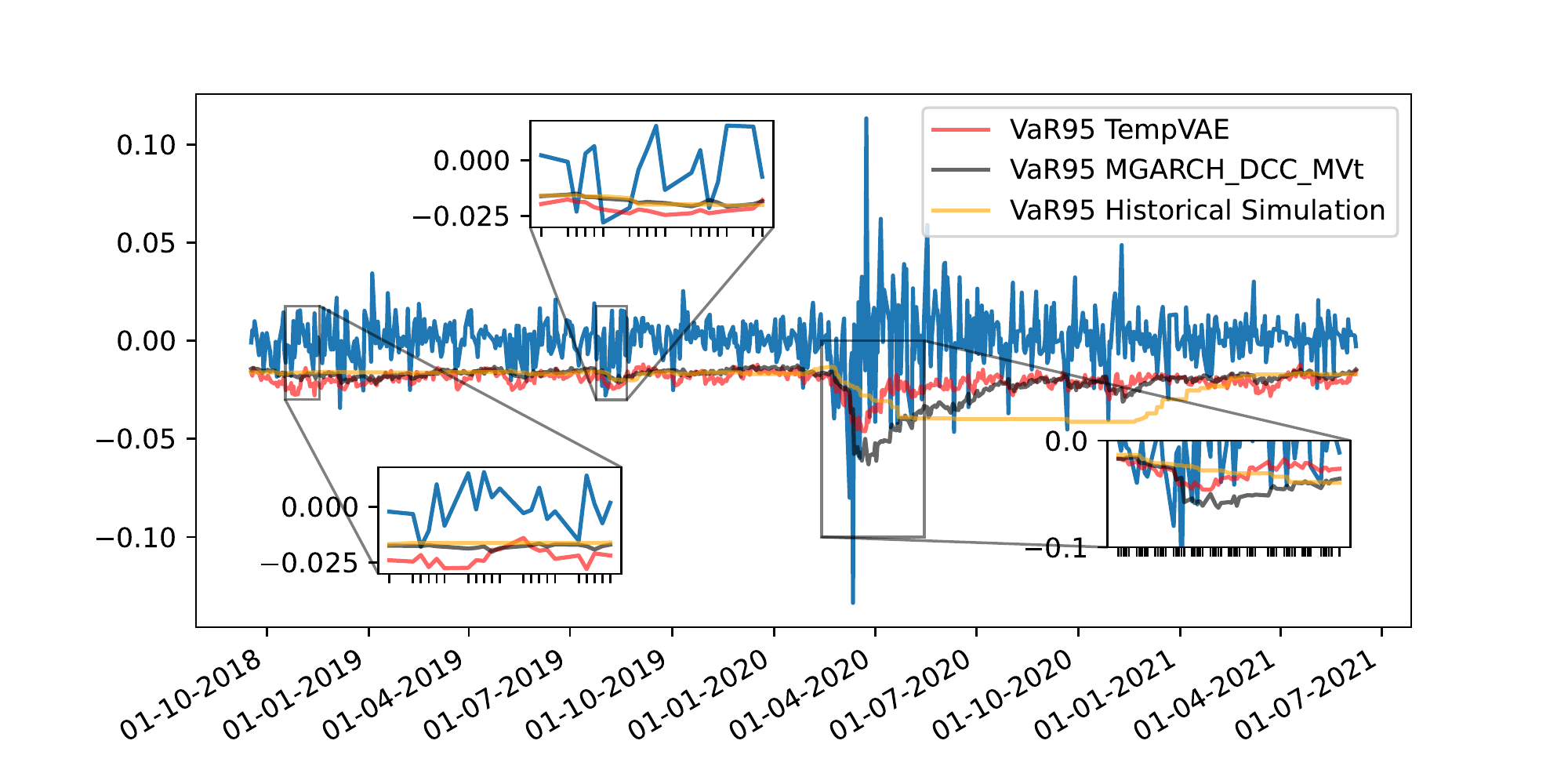}
		\caption{The VaR95 estimates for the two models TempVAE, DCC-GARCH-MVN and HS on a fraction of the test data. }
		\label{fig:VaR95 MGARCH,NSVM inset}
		
	\end{figure}
	
	\section{Conclusion}\label{sec: Conclusion}
	
	In this paper, we present a novel variational recurrent neural net with bottleneck structure known to autoencoders to handle sequential data. As we model the data on the basis of a time-dependent prior, we call the model temporal variational autoencoder (TempVAE). By ensuring that the information has to flow through the bottleneck, we leverage the auto-pruning property known to VAE. This way, we are able to find parsimonious model representations of the data, without using methods like grid search.
	
	We successfully apply the TempVAE to risk management. More precisely, we estimate the risk measure VaR and show that our model performs competitive to the considered benchmark models. We show that, expecting reasonable amount of VaR breaches, the model is performing best in terms of the considered RLF scores.
	
	Future research directions can incorporate applying the TempVAE architecture to other types of data sequences. Furthermore, the active dimensions in the bottleneck are not necessarily yielding disentangled representations. Research in this direction can foster the understanding of the auto-pruning and hence variational models further.
	
	\newpage
	\small

	\bibliography{TempVAE.bib}
	\newpage
	
	\normalsize\appendix
	
	\section{\texorpdfstring{$\beta$}{beta}-Annealing, Model Implementation and Data Preprocessing}\label{app: model implementation and beta annealing}
	
	\subsection{\texorpdfstring{$\beta$}{beta}-Annealing}
	
\begin{figure}[h!]
		\center
		\includegraphics[scale=0.65]{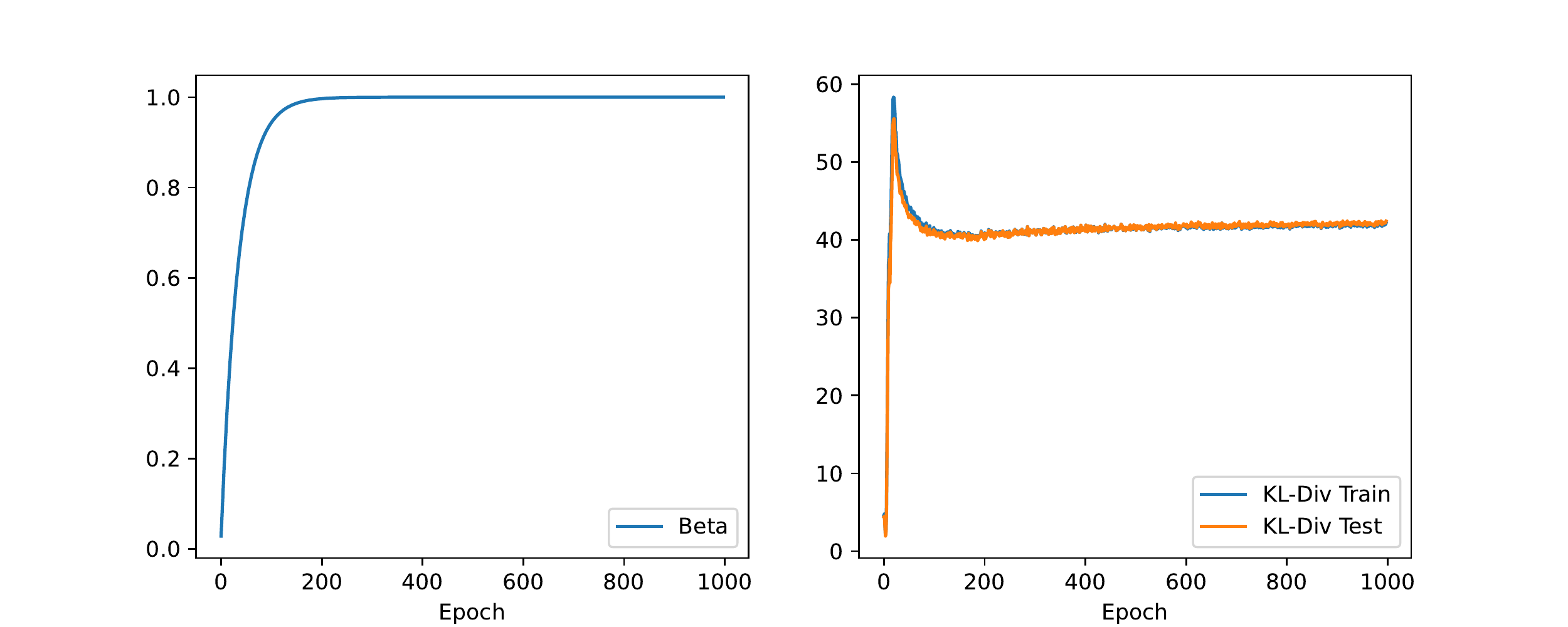}
		\caption{On the left: The KL influence is adjusted over $\beta$. During training we start with $\beta=0$ and increase it over time. We implement this by subtracting an exponentially decaying term from the ultimate $\beta$ value. We use a decay rate of $0.96$ and decay steps $20$. On the right: The evolution of the (negative) KL-Divergence over time. We can see how the values starts to decrease after the annealing has increased $\beta$ enough. }
		\label{fig:plot_betas_and_kl_div}
	\end{figure}

	The low signal-to-noise ratio in financial data makes it necessary to proceed delicately when modelling financial data. We propose to use annealing of the regularization to mitigate the initial effects of the auto-pruning. Specifically, we consider $\beta$-annealing of the KL-Divergence similar to \cite{Bowman2015}. We add $\beta\geq 0$ to the ELBO in \eqref{eq: Xu KL ELBO} to get
	\begin{align} \scriptstyle
	\ELBO_\beta(\btheta,\bphi) = \E_{q_{\bphi}(Z|R)}\left[\log p_{\btheta}(R|Z)\right] - \beta \cdot \sum_{t=1}^{T} \E_{q_{\bphi}(Z_{1:t-1}| R)} \Big[\KL\big(q_{\bphi}(Z_t| Z_{1:t-1},R) \: \big\lVert \: p_{\btheta}(Z_t| Z_{1:t-1}) \big)\Big]. \label{eq: Xu beta KL ELBO}
	\end{align}
	The KL influence is adjusted over $\beta$. During training we start with $\beta=0$ and increase it over time. In Figure \ref{fig:plot_betas_and_kl_div}, the annealing over 1000 epochs is depicted as well as the effect on the KL-Divergence is shown.
	
	\subsection{Model Implementation}
	
	\noindent We implement the model based on Section \ref{sec:The Temporal Variational Autoencoder}. For the RNN's we choose `Gated Recurrent Units' (see \cite{Cho2014}) with dimension 16 and for the MLP's we use normal Feedforward Networks that have two hidden layers with dimension 16 and the output layer with varying output dimension. For the latent variables we choose dimension 10. In our experimens, with this size of the latent dimension the model is overspecified and the auto-pruning removes not needed dimensions. 
	
	We initialize all MLP layer weights with the `Variance Scaling' initializer as proposed by \cite{He2015}. As activation function for the hidden layers we use the `Rectified Linear Unit' (ReLU). For the $\bmu^z$, $\bhmu^z$ and $\bmu^r$ layer we use no activation. We use a diagonal covariance structure for $\bSigma^z$ and $\bhSigma^z$ and a rank-1 perturbation output as given in \cite{Rezende2014} for $\bSigma^r$. 
	For all covariance layers, we use exponential activation for the diagonal entries. Further, we add L2 regularization with parameter $\lambda=0.01$ for the weights of the hidden layers of the MLP as well as dropout with a rate of $10\%$ to the RNN layers. In our experiments, both the dropout and the L2 regularization turned out to be crucial for the auto-pruning to work properly as well as for an adequate data modelling (see Appendix \ref{app:Regularization: L2, Dropout and KL-Divergence}). 
	
	In contrast to the model of \cite{Luo2018a}, we set the parameters of the prior (i.e. equations \eqref{eq:prior RNN MLP 1} - \eqref{eq:prior RNN MLP 3}) as non-trainable. Allowing a trainable prior results in an unfavourable performance regarding the signal identification as well as the risk-management application (see Appendix \ref{app:Analysing trainable prior}). In our point of view, the fixed prior serves as guidelines for the model to identify temporal transitions. If the prior is learned simultaneously with the encoder, past adaptions of the encoder to the temporal dynamics can become obsolete by changes in the prior parametrization.
	
	For the training of the model, we use the reparametrization trick as introduced by \cite{Kingma}. To approximate both expectations in \eqref{eq: Xu KL ELBO}, we use Monte Carlo with one sample.\footnote{We also experimented with a higher number of MC samples, but did not observe significant improvements. Therefore, we stick with the cheaper calculation.} Using the ADAM optimizer by \cite{Kingma2015}, we train the model for $1000$ epochs with a batch size of 256. We use an initial learning rate of $0.001$ with an exponential learning rate decay with decay rate $0.96$ and $500$ decay steps. If not explicitly excluded, we apply the $\beta$-annealing as described in Figure \ref{fig:plot_betas_and_kl_div} during training.\\

	\subsection{Data Preprocessing}
	
	To model the data with artificial neural networks, we split the data into a training and test set. Further do we use the empirical mean and standard deviation of the training part to standardize each ($i=1,\ldots,d$) univariate series $R_{1:T,i}$. Finally, we apply a sliding window on the $d$-dimensional series in \eqref{eq: d-dimensional data series} to generate $T-M$ intervals of size $M=21$. We define
	\[
	R_{t:t+M} := \left(R_{t},\ldots,R_{t+M}\right) \in \R^{d\times M},
	\]
	for $t= 1,\ldots,T-M$. After the preprocessing, we have different amounts of observations which we split into 66\% training and the rest test data:
	\begin{itemize}
		\item For the DAX data 5061 observations were split at $t=3340$ into 3340 training observations and 1721 test observations. 
		\item For the S\&P500 data 4872 observations were split at $t=3215$ into 3215 training observations and 1657 test observations. 
		\item For the noise data 5050 observations were split at $t=3333$ into 3333 training observations and 1717 test observations. 
		\item For each of the oscillating PCA data sets 9979 observations were split at $t=6586$ into 6586 training observations and 3393 test observations. 
	\end{itemize}
	
	\newpage
	\section{Ablation studies}\label{app:Further ablation studies}
	
	In this section, we validate some of the design choices for the TempVAE model. We will address each point in a separate section, starting of with the $\beta$-annealing.
	
	\subsection{Preventing Posterior Collapse with \texorpdfstring{$\beta$}{beta}-annealing}\label{sec: preventing posterior collapse with beta annealing}
	
	A deep neural net like the TempVAE has to be trained with caution to prevent posterior collapse. The $\beta$-annealing proposed in Appendix \ref{app: model implementation and beta annealing} is a useful tool to prevent this collapse. In this section, we compare the TempVAE with a model `TempVAE noAnneal' with no annealing, where we set $\beta=1$ instead of using $\beta$-annealing. Given the activity statistics of a model $A_{z}^{(m,k)}$, as described in \eqref{eq: time-dep activity} for $m=1,\ldots,M$ and $k=1,\ldots, \kappa$, we calculate the average count of active units as
	\begin{equation}\label{eq:avg count act}
	\bar{A}_z := \dfrac{1}{M \cdot \kappa} \sum_{m=1}^{M} \sum_{k=1}^{\kappa} \Ind_{\left\{A_{z}^{(m,k)} \geq 0.02\right\}}
	\end{equation}
	In Table \ref{tab: ablation anneal avg act}, we see the average count of active units for the two models and in Table \ref{tab: ablation anneal multN scores}, the NLL scores as calculated in Section \ref{sec: model fit and risk management} are reported.
	
	\begin{table}[htbp]
		\center
		\begin{tabular}{l|rrrrr}
			& DAX & Noise & Osc. PCA 2 & Osc. PCA 5 & Osc. PCA 10 \\ \hline
			TempVAE & 20\% & 0\% & 20\% & 20\% & 20\% \\ 
			TempVAE noAnneal & 0\% & 0\% & 10\% & 20\% & 10\% \\ 
		\end{tabular}
		\caption{The average count of active units given by \eqref{eq:avg count act} for the two models with and without annealing. We see that the model with the annealing is not decreasing the number of average active dimensions when increasing the number of latent signals. For the model without annealing this is not the case.}
		\label{tab: ablation anneal avg act}
	\end{table}

	\begin{table}[htbp]
		\center
		\begin{tabular}{l|rrrrr}
			& DAX & Noise & Osc. PCA 2 & Osc. PCA 5 & Osc. PCA 10 \\ \hline
			TempVAE & \textbf{20.29} & 31.48 & \textbf{-38.51} & \textbf{-1.40} & \textbf{11.07} \\ 
			TempVAE noAnneal & 27.49 & 31.44 & 3.58 & 2.60 & 13.28 \\ 
		\end{tabular}
		\caption{The negative log-likelihood for the two models on the test set. We see the TempVAE is outperforming the version without annealing on every relevant data set. Only for the noise case, where there is nothing to learn, the results are comparable. }
		\label{tab: ablation anneal multN scores}
	\end{table}
	
	\noindent Using the annealing, a latent process in financial data is identified. Without annealing, a posterior collapse occurs and no encoding information is propagated to the decoder. Notice that the amount of active dimensions need not match the amount of latent signals. Because of non-linearities it can be possible for the model to use less dimensions than needed to adequately model the data (see Appendix \ref{app:TempVAE latent space act osc pca 5 10}).

	\subsection{Comparison to a Model with Trainable Prior Parameters}\label{app:Analysing trainable prior}
	
\begin{figure}[h!]
		\center
		\includegraphics[scale=0.55]{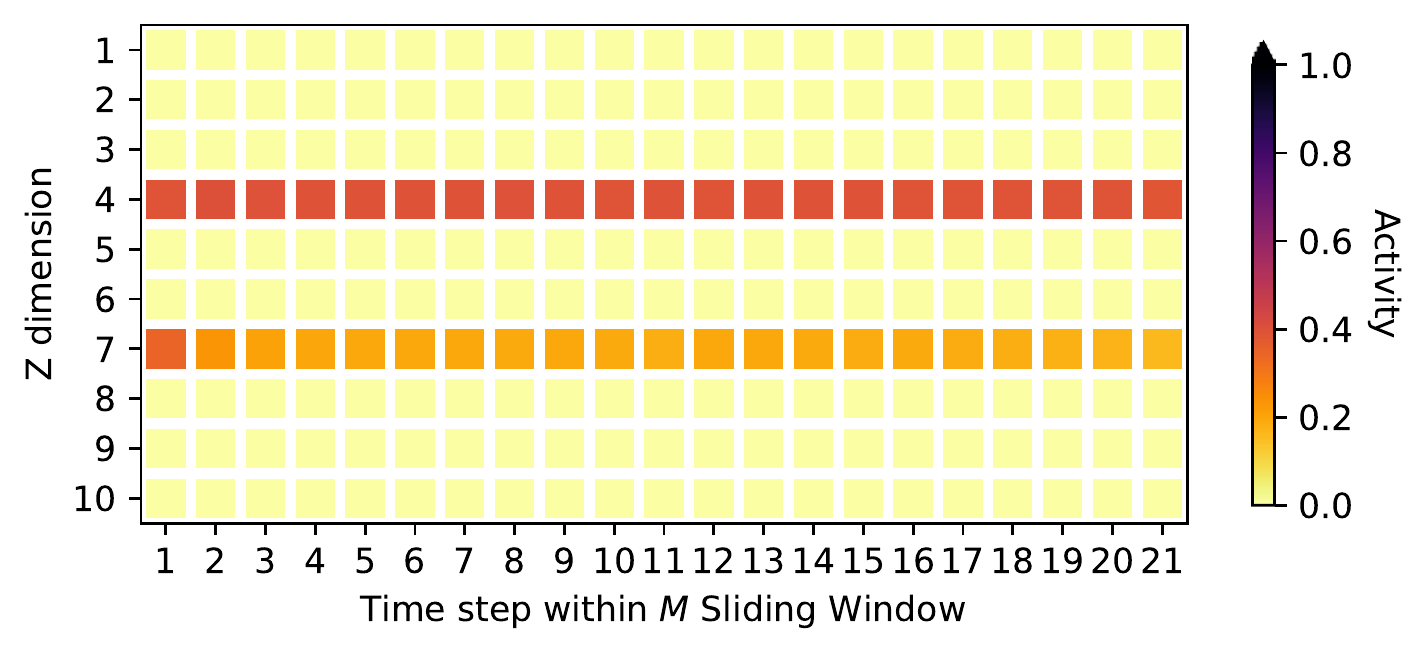}
		\caption{The activity statistics for the ``DAX'' data for the model `TempVAE trainPrior'. We consider sequences of size $M=21$ as input. Depicted are the activity values of statistic \eqref{eq: time-dep activity} for the $\kappa=10$ dimensional latent space. }
		\label{fig:plot_Activities_heatmap_dax_model187}
	\end{figure}

\begin{figure}[h!]
		\center
		\includegraphics[scale=0.55]{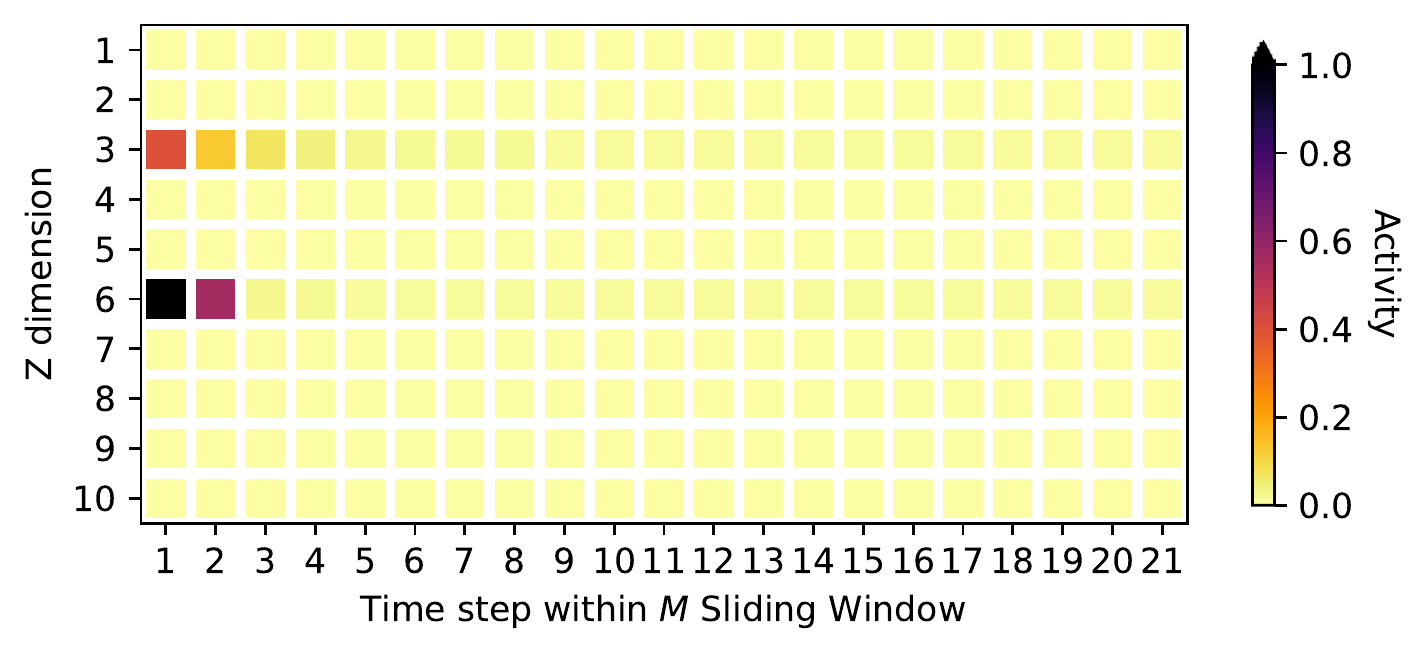}
		\caption{The activity statistics for the ``oscillating PCA'' data for the model `TempVAE trainPrior'. We consider sequences of size $M=21$ as input. Depicted are the activity values of statistic \eqref{eq: time-dep activity} for the $\kappa=10$ dimensional latent space. }
		\label{fig:plot_Activities_heatmap_oscpca_model187}
	\end{figure}

	Allowing a trainable prior distribution $p_{\btheta}(Z)$ yields unfavourable performance regarding the auto-pruning as well as a worse fit to the data. In this section, we find that using a non-trainable prior $p_{\btheta}(Z)$ yields better results in terms of the auto-pruning functionality as well as of the fit of the model. To see this, we compare the TempVAE model with a version `TempVAE trainPrior', where the prior is trainable. Figure \ref{fig:plot_Activities_heatmap_dax_model187} and \ref{fig:plot_Activities_heatmap_oscpca_model187} display the activities of the `TempVAE trainPrior' bottleneck for the DAX data and for the Oscillating PCA data respectively.
	
\begin{figure}[h!]
		\center
		\includegraphics[scale=0.4]{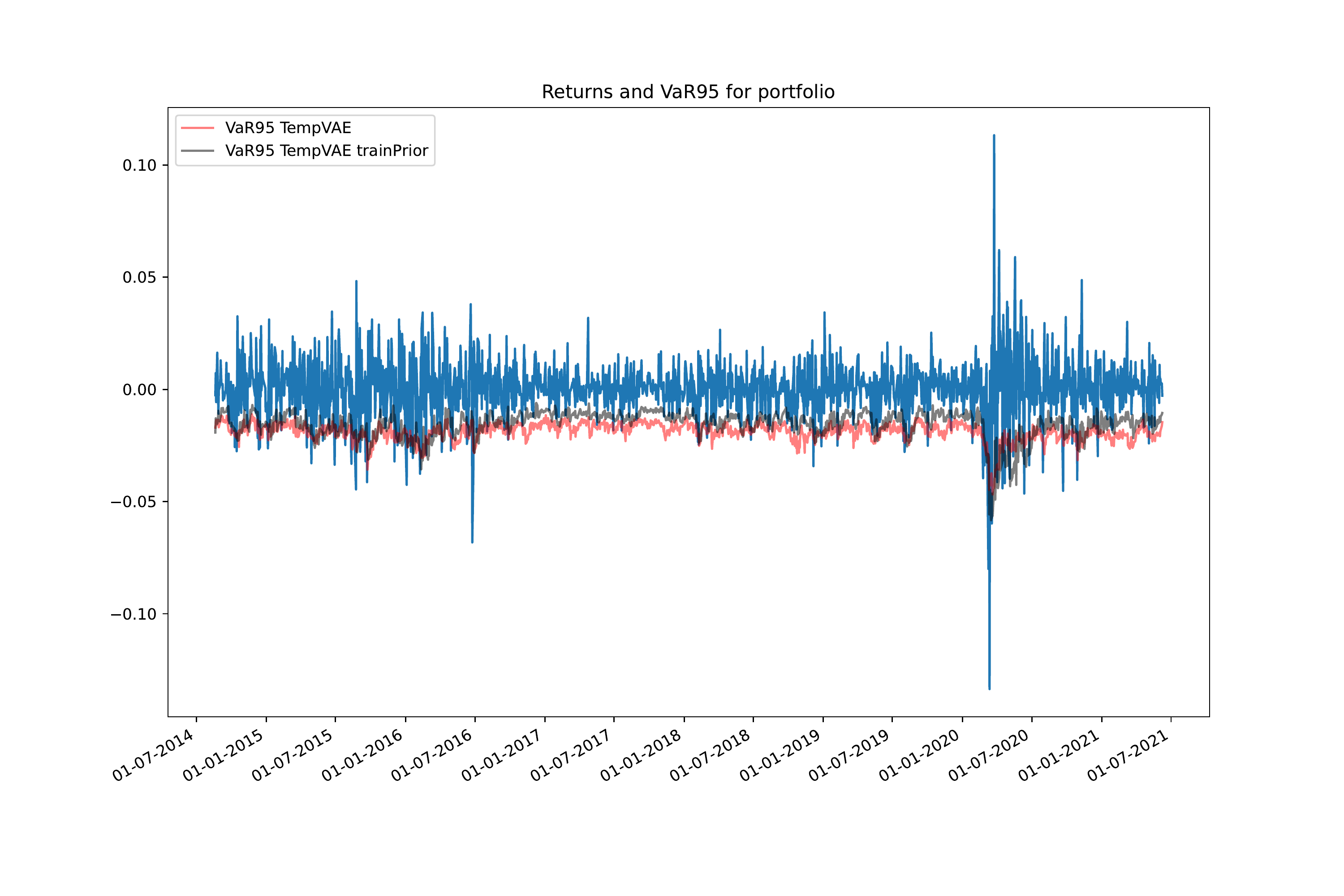}
		\caption{VaR95 forecasts of both models `TempVAE' and `TempVAE trainPrior' on the test set. The VaR forecasts for the model with trainable prior are more volatile. }
		\label{fig: ablation VaR95 trainable Prior}
	\end{figure}

	\noindent Comparing these figures with Figure \ref{fig:plot_Activities_heatmap_osc_pca}, we notice that the value of the activity statistics drastically decreased. For the model with trainable prior, it is questionable if any information from the input is propagated to the output for the `Oscillating PCA' data at later time steps. \\
	\\
	Comparing the VaR forecasts of both models in Figure \ref{fig: ablation VaR95 trainable Prior}, we  see that the VaR forecasts for the model with trainable prior are more volatile. This also affects the average amount of breaches of this estimator as can be seen in Table \ref{tab: ablation VaR95 breches trainable prior}. The TempVAE outperforms the version with trainable prior, as the amount of breaches for the VaR95 should be close to 5 and for VaR99 to 1. In our point of view, setting the prior as non-trainable gives the model more stability during training. This results in a more consistent output e.g. for the VaR estimates.
	
	\begin{table}[htbp]
		\center
		\begin{tabular}{l|rr}
			Model & Br95 & Br99 \\ \hline
			TempVAE & \textbf{4.8} & \textbf{1.3} \\ 
			TempVAE trainPrior & 8.2 & 3.1 \\ 
		\end{tabular}
		\caption{The count of average breaches for the model TempVAE and the version with trainable parameters for the prior $p_{\btheta}(Z)$. TempVAE outperforms the trainable prior version. For the average breaches, the respective quantile assumption is the optimal value. Hence for VaR95 the best value is 5 and for VaR99 the best value is 1.}
		\label{tab: ablation VaR95 breches trainable prior}
	\end{table}
	
	\subsection{Auto-Regressive Structure for the Observables Distribution}
	
	Modelling an auto-regressive connection for the observations produces questionable results due to over-fitting. In this section, we compare the model TempVAE with a the model `TempVAE AR', a version of the model with auto-regressive structures as proposed, e.g. by \cite{Chen2020} or \cite{Luo2018a}. The implementation of the model `TempVAE AR' is identical to the model TempVAE except for the generative distribution. Instead of \eqref{eq:generative RNN MLP 1}-\eqref{eq:generative RNN MLP 3}, we implement this part as
	\begin{align}
	\left\{\bmu_t^r,\bSigma_t^r\right\} &= \mathrm{MLP}^r_G\left(h_t^r\right), \label{eq:ablation AR generative RNN MLP 1}\\
	h_t^r &= \mathrm{RNN}_G^r\left(h_{t-1}^r,z_{t},r_{t-1}\right),\label{eq:ablation AR generative RNN MLP 2}\\
	r_{t} &\sim \Nor\left(\bmu_t^r,\bSigma_t^r\right)\label{eq:ablation AR generative RNN MLP 3}.
	\end{align}
	Therefore, past observations are influencing the current distributions. If such a dependency structure is assumed given a sequence of observations $R_{1:T}$, the latent variables become dependent on the whole observed sequence, as argued in \cite{Luo2018a}. As we account for this already in our proposed model for the encoder (see \eqref{eq:encoder RNN MLP 1}-\eqref{eq:encoder RNN MLP 5}), we do not have to adapt for this part. 
	
	\noindent Looking at Table \ref{tab: ablation study tempvae ar}, we see that TempVAE outperforms `TempVAE AR' on all test sets. In fact, we can observe overfitting for the `TempVAE AR' model whereas the TempVAE did not suffer from this. Common VAE possess an inherent robustness against overfitting (see e.g. \cite{Kingma2014}) originating from the regularization with the KL-Divergence. As the structure of the TempVAE enforces this regularized bottleneck structure, we are not surprised that we do not observe overfitting with the TempVAE. On the other side, `TempVAE AR' has the possibility to ignore the regularized bottleneck and hence overfitting becomes more of a concern.
	
	\begin{table}[htbp]
		\center
		\begin{tabular}{l|rrrrr}
			& DAX & Noise & Osc. PCA (2) & Osc. PCA (5) & Osc. PCA (10) \\ \hline
			TempVAE & \textbf{20.29} & \textbf{31.48} & \textbf{-38.51} & \textbf{-1.40} & \textbf{11.07} \\ 
			TempVAE AR & 22.68 & 32.91 & -34.72 & 25.33 & 69.26 \\ 
		\end{tabular}
		\caption{The negative log-likelihood for the two models TempVAE and `TempVAE AR' on the test set. We see the TempVAE is outperforming the version with auto-regressive structure, even in the case of the Noise data set. This is because the `TempVAE AR' model was overfitting this data. }
		\label{tab: ablation study tempvae ar}
	\end{table}
	
	\noindent Apart from these results, another point speaks in favor of the TempVAE without auto-regressive structure for the observations. As information does not necessarily has to pass the bottleneck, analyzing the latent space activity becomes tedious. Even for a completely collapsed posterior $q$ the model can use information from the input sequence to model the output.
	
	\vspace{1cm}
	\subsection{Using a Diagonal Covariance Matrix \texorpdfstring{$\bSigma^r_t$}{Sigma}}\label{app: diag cov}
	
	Here, we compare the `TempVAE' with the model `TempVAE diag', a version of the model, where the output covariance matrix for the observables is modelled as diagonal and not via the rank-1-perturbation as given in Appendix \ref{app: model implementation and beta annealing}. As we can see in Table \ref{tab: ablation diag multN scores}, the average amount of VaR breaches is approximately the same for the two model architectures. This indicates that by the usage of the latent variables the covariance structure of the data can be explained well enough to get a comparable performance for the VaR estimates. We still decided to propose the rank-1 perturbation for the model, as \cite{Luo2018a} were able to improve their fit this way.
	
	\begin{table}[htbp]
		\center
		\begin{tabular}{l|rr}
			Model & Br95 & Br99 \\ \hline
			TempVAE  & 4.8 & 1.3 \\ 
			TempVAE diag & 5.3 & 1.6 \\ 
		\end{tabular}
		\caption{The count of average breaches for the model TempVAE and the version with diagonal covariance matrix for $\bSigma^r_t$. For the average breaches, the respective quantile assumption is the optimal value. Hence for VaR95 the best value is 5 and for VaR99 the best value is 1.}
		\label{tab: ablation diag multN scores}
	\end{table}
	
	\newpage
	\subsection{Setting \texorpdfstring{$\bmu^r_t \equiv0$}{mean zero}}\label{app: mu zero}
	
\begin{figure}[h!]
		\center
		\includegraphics[scale=0.55]{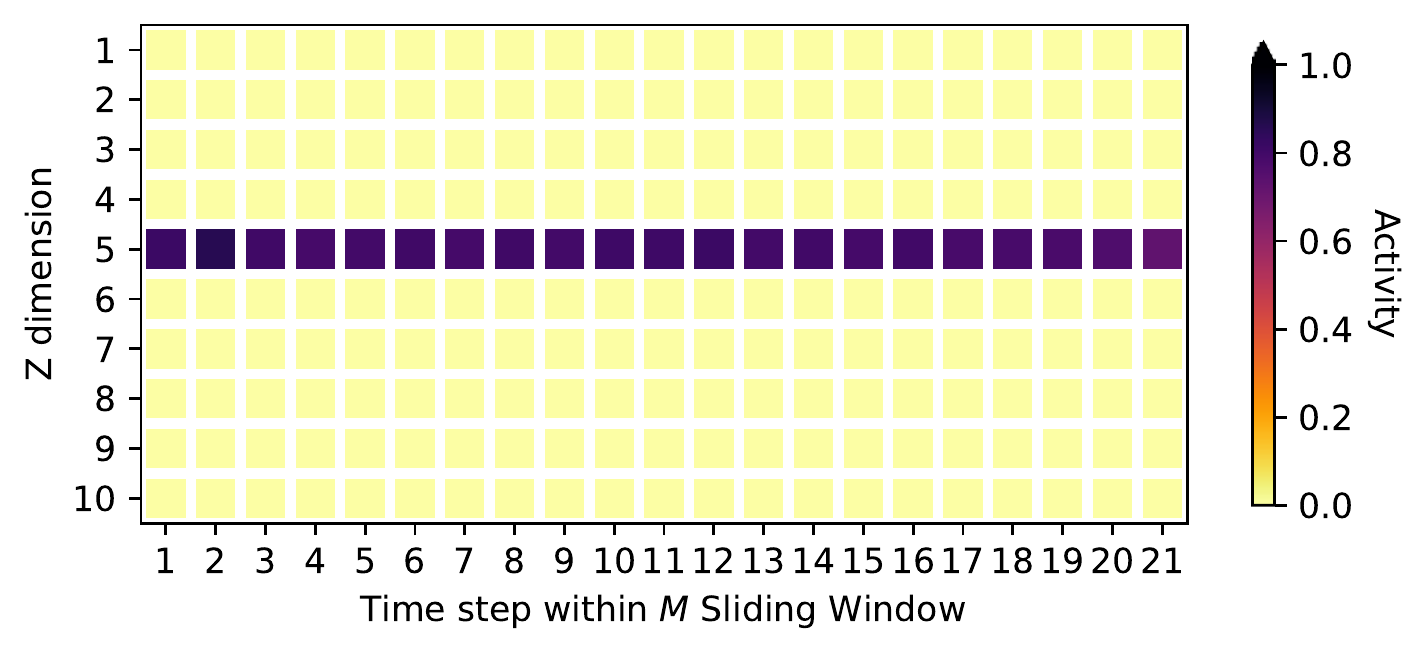}
		\caption{The activity statistics for the ``DAX'' data for the model `TempVAE zeroMean'. We consider sequences of size $M=21$ as input. Depicted are the activity values of statistic \eqref{eq: time-dep activity} for the $\kappa=10$ dimensional latent space. }
		\label{fig:plot_Activities_heatmap_dax_model194}
		
	\end{figure}

\begin{figure}[h!]
		\center
		\includegraphics[scale=0.55]{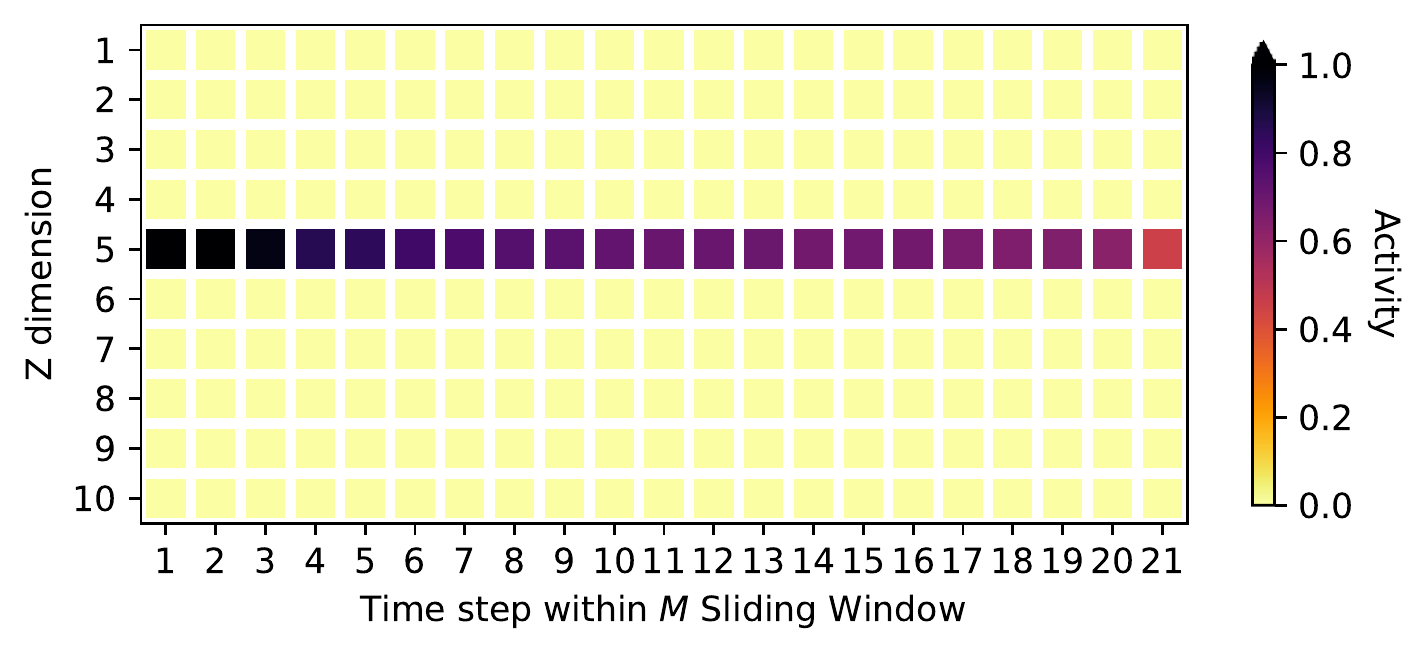}
		\caption{The activity statistics for the ``Oscillating PCA 2'' data for the model `TempVAE zeroMean'. We consider sequences of size $M=21$ as input. Depicted are the activity values of statistic \eqref{eq: time-dep activity} for the $\kappa=10$ dimensional latent space. }
		\label{fig:plot_Activities_heatmap_oscpca_model194}
		
	\end{figure}

	In their paper, \cite{Xu2021} motivate to model only the covariance $\bSigma^r_t$ of the observable distribution \eqref{eq:assumption r}. We therefore compare the TempVAE with the model `TempVAE zeroMean', the same model but with the mean parameter fixed to an output of zero. Figure \ref{fig:plot_Activities_heatmap_dax_model194} and \ref{fig:plot_Activities_heatmap_oscpca_model194} show the activity statistics for the ``DAX'' and ``Oscillating PCA 2'' datasets.
	
\begin{figure}
		\center
		\includegraphics[scale=0.5]{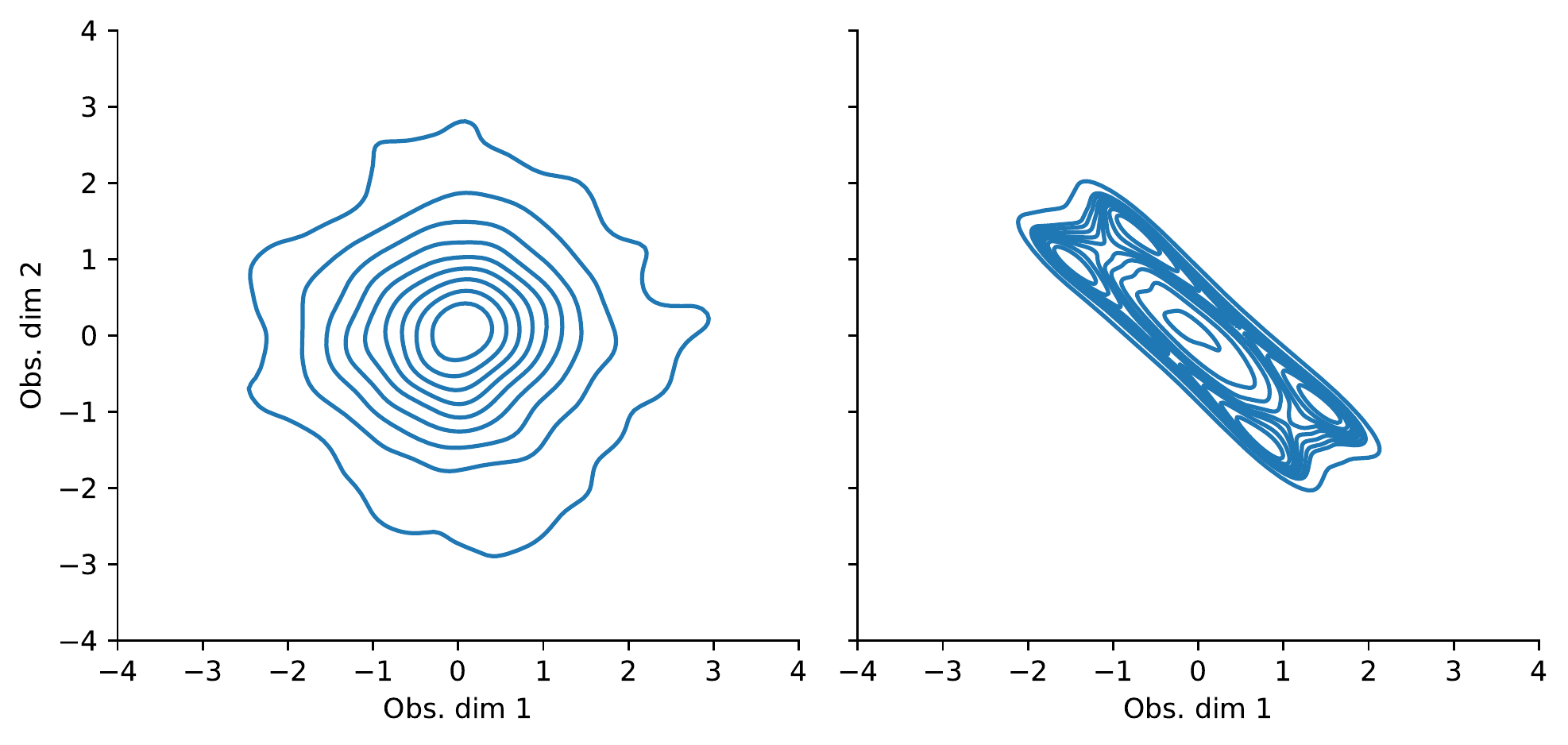}
		\caption{The kernel density estimated distribution of the ``oscillating PCA 2'' data. On the left we see the model `TempVAE zeroMean' and on the right the historical data. }
		\label{fig:plot_scatter_osc_pca_model194}
		
	\end{figure}

	\noindent Here, we can observe that the number of identified active nodes per timestep is less in comparison with the TempVAE model. A smaller latent space occupation than given by the amount of latent signals is possible due to non-linearities of the net. But in our point of view, it is rather questionable that one latent dimension suffices to capture the signal in both the ``Oscillating PCA 2''  dataset and the ``DAX'' dataset. Further, comparing the kernel density estimated distributions of the ``Oscillating PCA 2''  dataset in Figure \ref{fig:plot_scatter_osc_pca_model194} and \ref{fig:plot_scatter_osc_pca}, we see that the model is not able to correctly model the two dimensions of the signal inherent in the data. Therefore, we decide to model the mean parameter.

	\newpage
	\subsection{Regularization: L2, Dropout and KL-Divergence}\label{app:Regularization: L2, Dropout and KL-Divergence}
	
	In this section, we compare the TempVAE to five other models where different regularization procedures are set inactive. The models we consider are
	\begin{enumerate}
		\item `TempVAE'
		\item `TempVAE det': The KL-Divergence is switched off and the bottleneck uses only a mean parameter whereas the covariance is set to zero. Therefore, the bottleneck is deterministic and the auto-pruning switched off.
		\item `TempVAE no dropout/L2': Dropout and L2 regularization are switched off.
		\item `TempVAE no L2': L2 regularization is switched off.
		\item `TempVAE no dropout': Dropout is switched off.
		\item `TempVAE det no dropout/L2': `TempVAE det' and `TempVAE no dropout/L2' combined.
	\end{enumerate}
	
	\noindent In Table \ref{tab: ablation study regularizations (dropout,l2,kl-div)}, the average count of active dimensions is shown. As we can see, only the model `TempVAE' is providing favourable results. The deterministic models show no auto-pruning for the Oscillating PCA data versions while the model `TempVAE det no dropout/L2' is even activating most of the latent nodes when modeling the Noise data. This extensive use of bottleneck nodes is a sign of overfitting. It is important to understand that if the KL-Divergence is switched off for these models, the reconstruction error, that is, the first term in equation \eqref{eq: Xu beta KL ELBO}, is the only factor driving the encoder model training. Hence, the encoder aims to provide latent variables that are best suited for reconstruction just as in a normal autoencoder model. Since encoder and prior distribution are disconnected without the KL-Divergence term, the forecast of these models is highly doubtful. Nonetheless the reconstruction has been learned and therefore we can analyze the latent space activity.
	
	The stochastic models without L2, dropout or both also show questionable results. Therefore, we conclude that both regularizations are essential to learn the deep neural net and for the auto-pruning to work properly.

	\begin{table}[htbp]
		\center
		\begin{tabular}{l|rrrrr}
			Model & dax & noise & oscPCA2 & oscPCA5 & oscPCA10 \\ \hline
			TempVAE & 20\% & 0\% & 20\% & 20\% & 20\% \\ 
			TempVAE det & 0\% & 0\% & 90\% & 81\% & 90\% \\ 
			TempVAE no dropout/L2 & 20\% & 0\% & 31\% & 76\% & 30\% \\ 
			TempVAE no L2 & 20\% & 6\% & 20\% & 20\% & 30\% \\ 
			TempVAE no dropout & 20\% & 0\% & 31\% & 41\% & 40\% \\ 
			TempVAE det no dropout/L2 & 90\% & 90\% & 90\% & 80\% & 90\% \\ 
		\end{tabular}
		\caption{The average count of active units given by \eqref{eq:avg count act} for the models, where different regularizations are excluded. }
		\label{tab: ablation study regularizations (dropout,l2,kl-div)}
	\end{table}

\begin{figure}
		\center
		\includegraphics[scale=0.5]{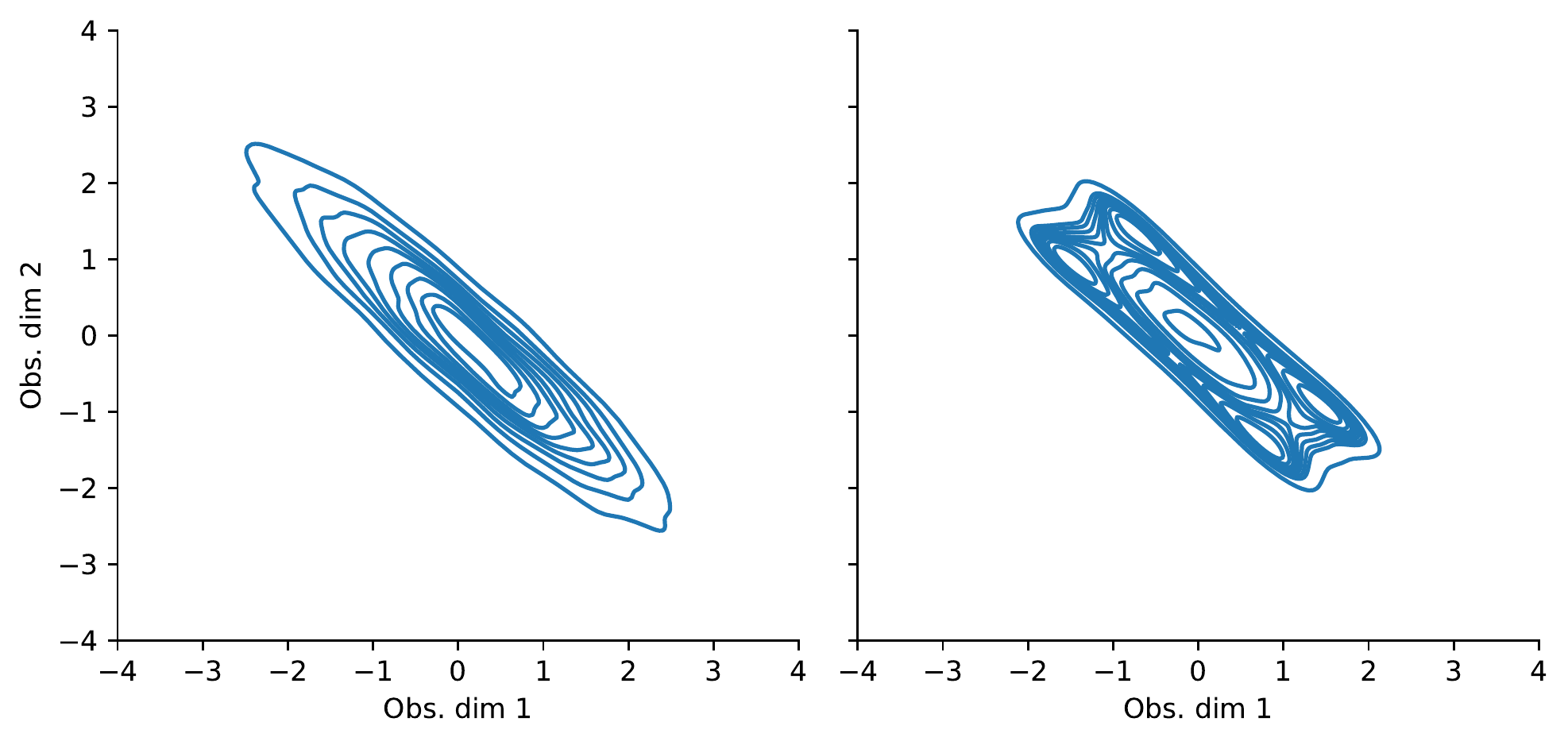}
		\caption{The kernel density estimated distribution of the ``oscillating PCA 2'' data for the model  `TempVAE no dropout'. On the left we see the `TempVAE no dropout' model and on the right the historical data. }
		\label{fig:plot_scatter_osc_pca_model193}
		
	\end{figure}

	Furthermore, we found that none of the models without dropout were able to adequately model the data structure of the ``Oscillating PCA 2''. Figure \ref{fig:plot_scatter_osc_pca_model193}, displays the kernel density estimated distribution of the first two dimensions of the data ``Oscillating PCA 2''. In contrast to Figure \ref{fig:plot_scatter_osc_pca}, the plot for the model `TempVAE' the structure is poorly fit.

	\subsection{Encoder Dependency}\label{app:Encoder depencency}
	
	In this section, we we compare the `TempVAE' model with `TempVAE backwards', a version where the encoder structure is not implemented as in \eqref{eq:encoder RNN MLP 1}- \eqref{eq:encoder RNN MLP 5}, but with a backward RNN, given by

	\begin{align}
	\left\{\bhmu_t^z,\bhSigma_t^z\right\} &= MLP^z_I\left(\hat{h}_t^z\right)\\
	\hat{h}_t^z &= RNN_I^z\left(\hat{h}_{t-1}^z,z_{t-1},\hat{h}_t^{\leftarrow}\right)\\
	\hat{h}_t^{\leftarrow} &= RNN_I^z\left(\hat{h}^{\leftarrow}_{t+1},r_{t+1}\right)\\
	z_{t} &\sim \Nor\left(\bhmu_t^z,\bhSigma_t^z\right).
	\end{align}
	This is in consensus with the dependency structure of the posterior $p_{\btheta}(Z|R)$, shown in \eqref{eq:p posterior z|r future dependency}.

	\begin{figure}
		\center
		\includegraphics[scale=0.5]{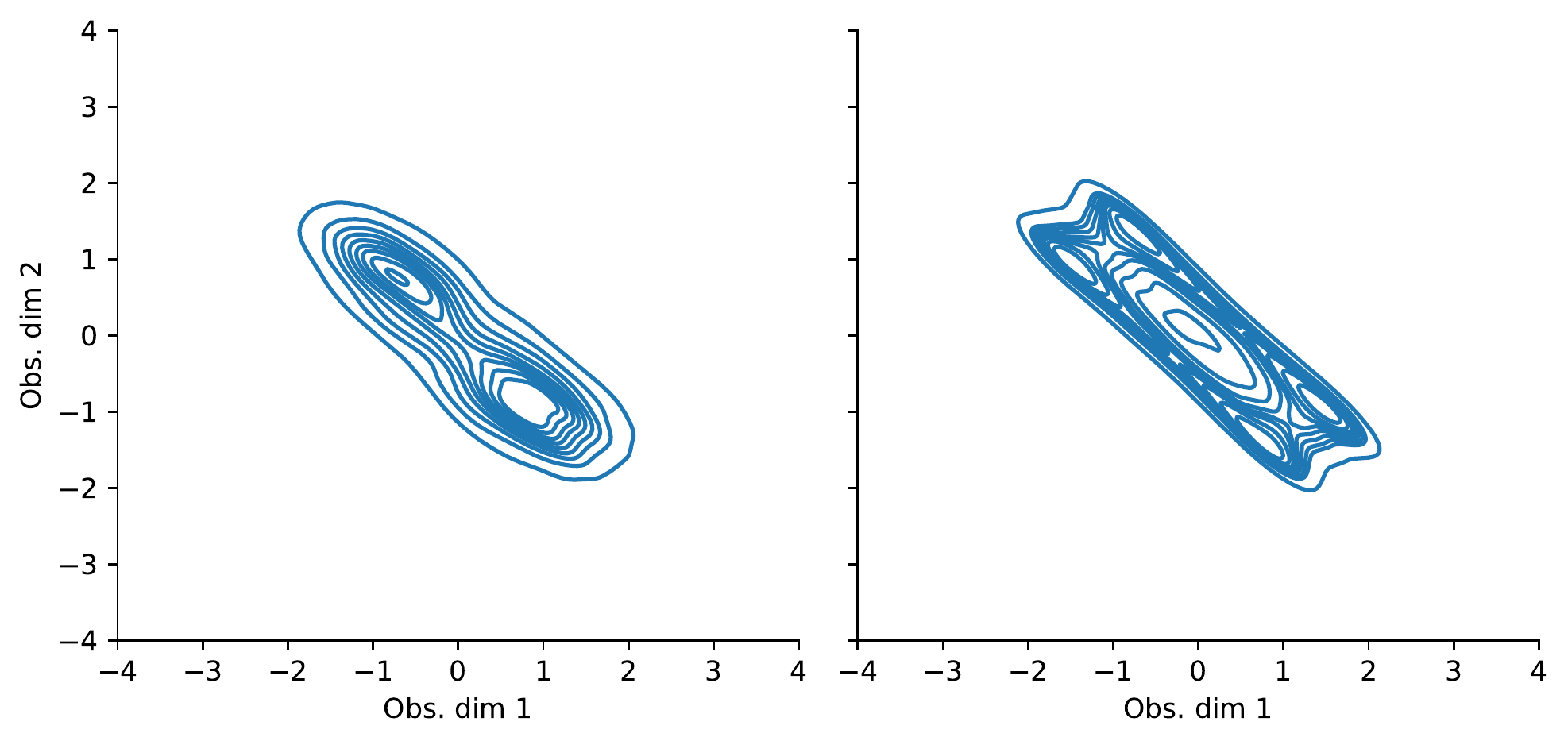}
		\caption{The kernel density estimated distribution of the ``oscillating PCA 2'' data for the model  `TempVAE backwards'. On the left we see the `TempVAE backwards' model and on the right the historical data.}
		\label{fig:plot_scatter_osc_pca_model186}
		
	\end{figure}

	Figure \ref{fig:plot_scatter_osc_pca_model186} displays the KDE estimation of the first two dimensions of the oscillating PCA data. In contrast to Figure \ref{fig:plot_scatter_osc_pca}, the plot for the model `TempVAE' the structure is poorly fit.

	\begin{table}[htbp]
		\center
		\begin{tabular}{l|rrrrr}
			Model & dax & noise & oscPCA2 & oscPCA5 & oscPCA10 \\ \hline
			TempVAE & 20\% & 0\% & 20\% & 20\% & 20\% \\ 
			TempVAE backwards & 10\% & 0\% & 10\% & 0\% & 0\% \\ 
		\end{tabular}
		\caption{The average count of active units given by \eqref{eq:avg count act} for the two models with different encoder architecture. }
		\label{tab: ablation backwards encoder avg act}
	\end{table}
	
	Table \ref{tab: ablation backwards encoder avg act} shows that the auto-pruning is performing poorly when using only the backwards RNN for the model. As the encoding model is only an variational approximation to the actual posterior, we therefore decide to use the bidirectional architecture for our model.

	\section{GARCH and DCC-GARCH}\label{app:GARCH and MGARCH} 
	
	As benchmarks for the fit of the financial data, we consider three models. 
	\begin{enumerate}
		\item GARCH: A GARCH(1,1) model, introduced by \cite{Bollerslev1986}.
		\item DCC-GARCH-MVN: A Dynamic Conditional Correlation GARCH(1,1), introduced by \cite{Engle2012}, with a multivariate Gaussian distribution assumption for the error term.
		\item DCC-GARCH-MVt: A Dynamic Conditional Correlation GARCH(1,1), with a multivariate t distribution assumption for the error term.
	\end{enumerate} 
	As the GARCH model is univariate, we model each dimension of our observed data separately. Therefore, we assume a correlation of 0 among the observed assets. The DCC-GARCH models can be seen a a multivariate extension of the first, where correlations are modeled as well.
	All models are linear and optimized by using Maximum Likelihood. We used the `rmgarch' library in R (see \cite{Ghalanos2019}).

	\section{Activities on the Oscillating PCA Data Sets}\label{app:TempVAE latent space act osc pca 5 10}

	\begin{figure}[h!]
		\center
		\includegraphics[scale=0.55]{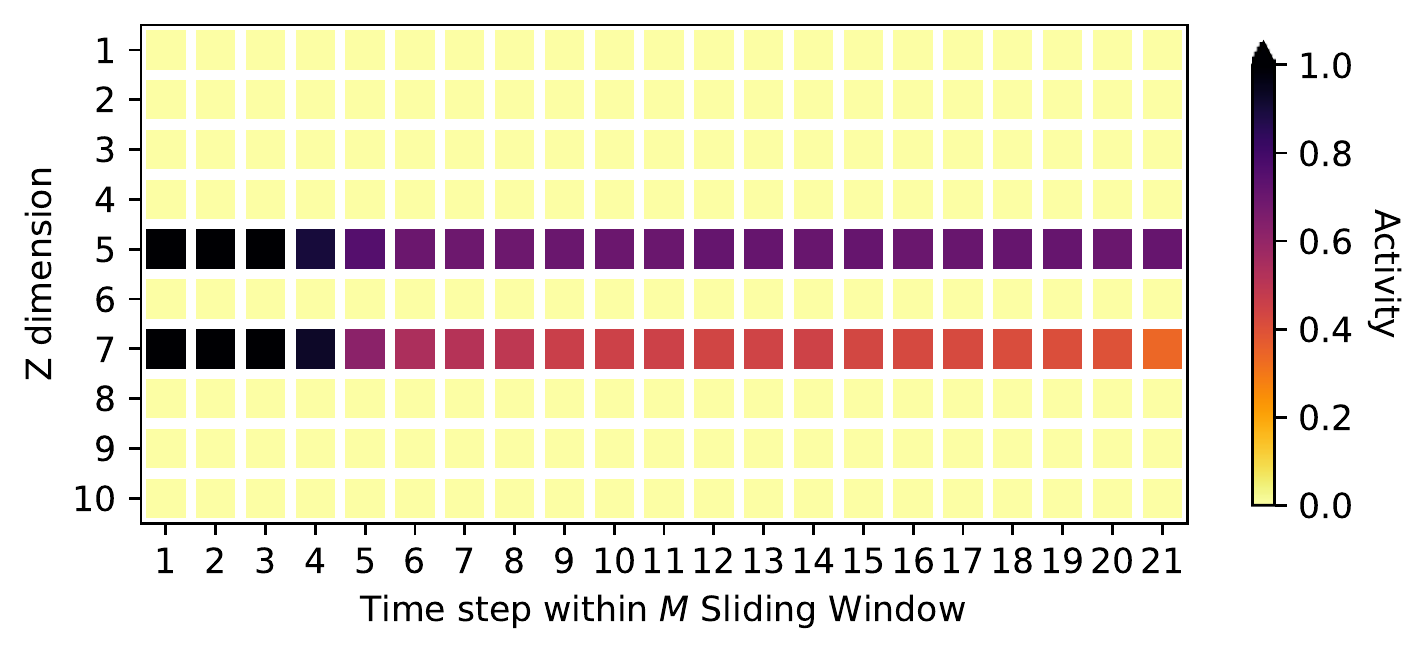}
		\caption{The activity statistics for the ``oscillating PCA 5'' data. We consider sequences of size $M=21$ as input. Depicted are the activity values of statistic \eqref{eq: time-dep activity} for the $\kappa=10$ dimensional latent space. }
		\label{fig:plot_Activities_heatmap_osc_pca5}
		
	\end{figure}
	
	\begin{figure}[h!]
		\center
		\includegraphics[scale=0.55]{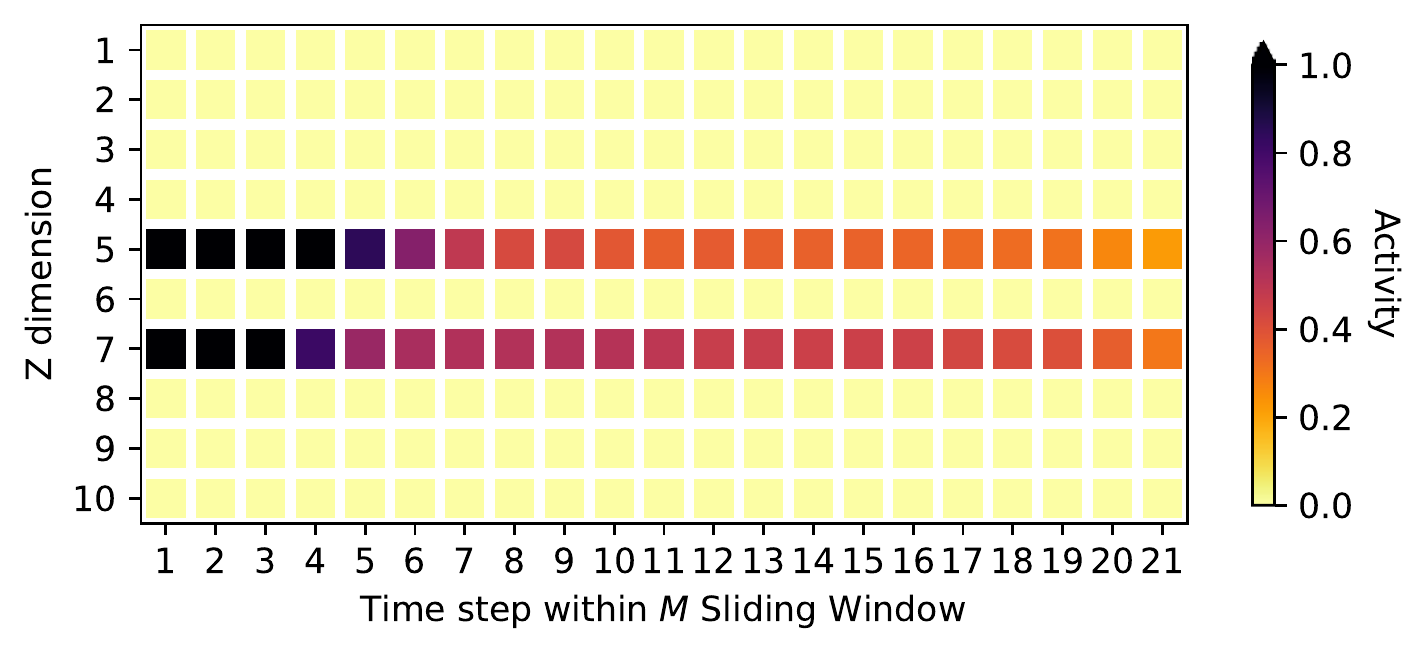}
		\caption{The activity statistics for the ``oscillating PCA 10'' data. We consider sequences of size $M=21$ as input. Depicted are the activity values of statistic \eqref{eq: time-dep activity} for the $\kappa=10$ dimensional latent space. }
		\label{fig:plot_Activities_heatmap_osc_pca10}
		
	\end{figure}
	
\begin{figure}
		\center
		\includegraphics[scale=0.5]{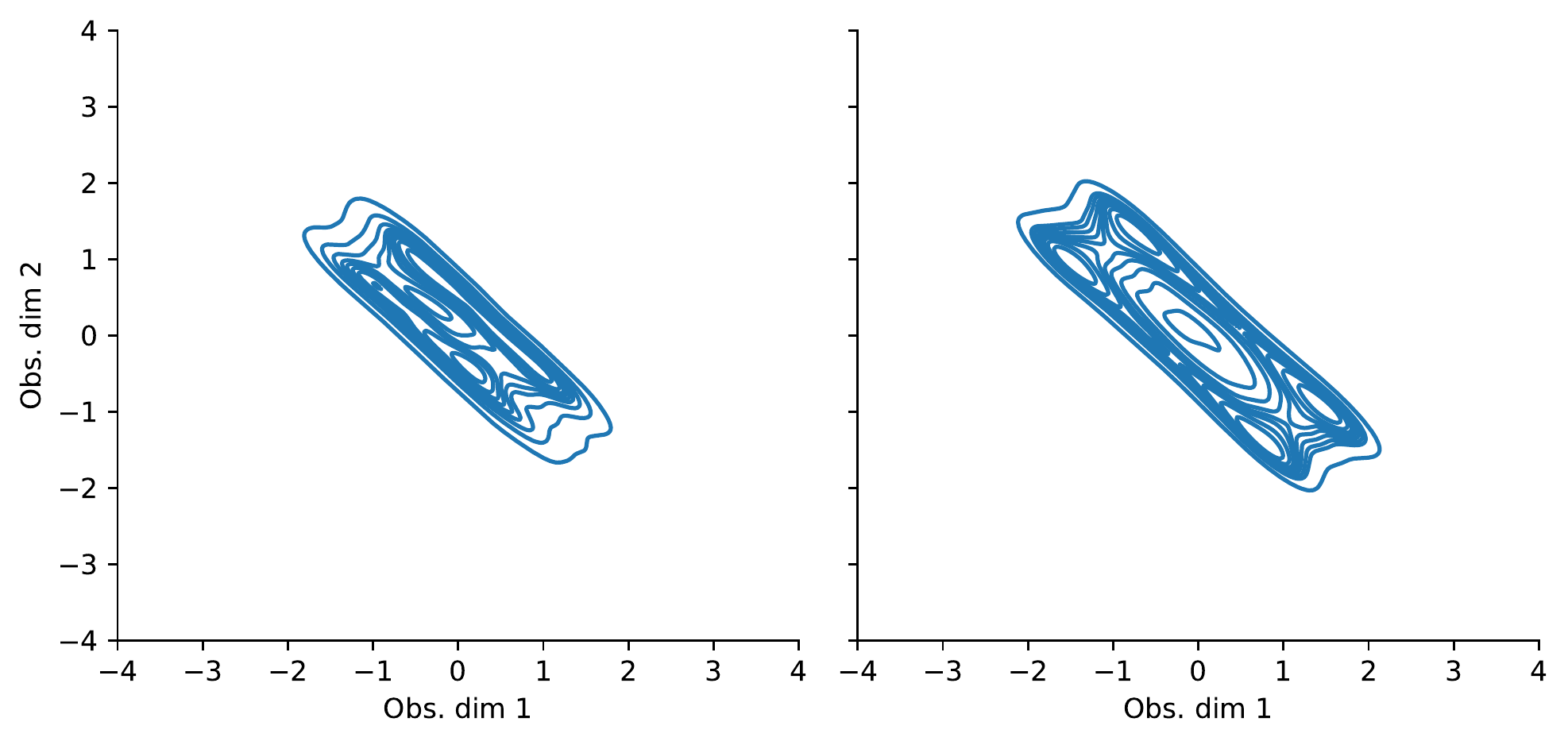}
		\caption{The kernel density estimated distribution of the first two dimensions $R_{t,1:2}$ of the ``oscillating PCA 2'' data. On the left we see the TempVAE model and on the right the historical data. }
		\label{fig:plot_scatter_osc_pca}
		
	\end{figure}

\begin{figure}
		\center
		\includegraphics[scale=0.5]{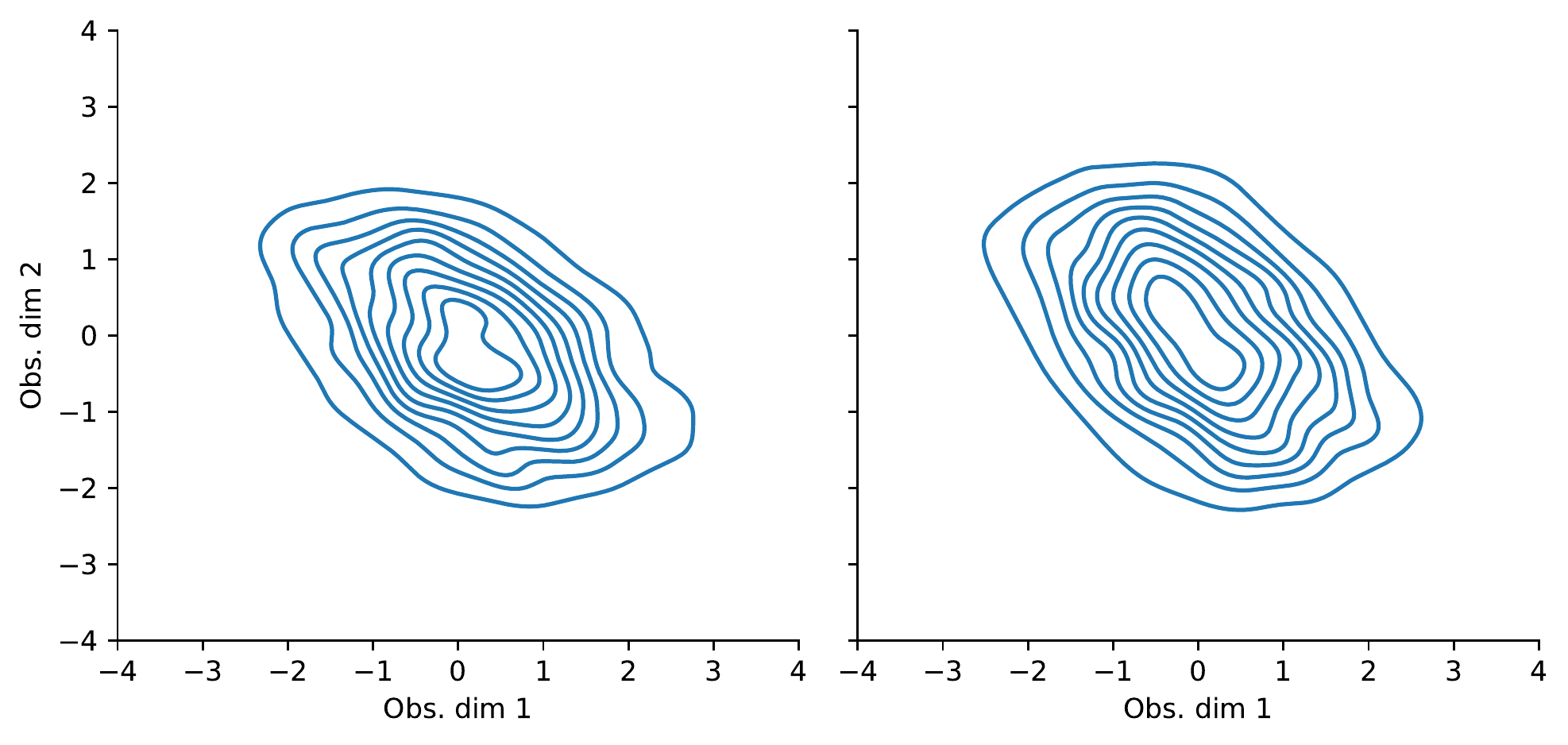}
		\caption{The kernel density estimated distribution of the first two dimensions $R_{t,1:2}$ of the ``oscillating PCA 5'' data. On the left we see the TempVAE model and on the right the historical data. }
		\label{fig:plot_scatter_osc_pca5}
		
	\end{figure}

\begin{figure}
		\center
		\includegraphics[scale=0.5]{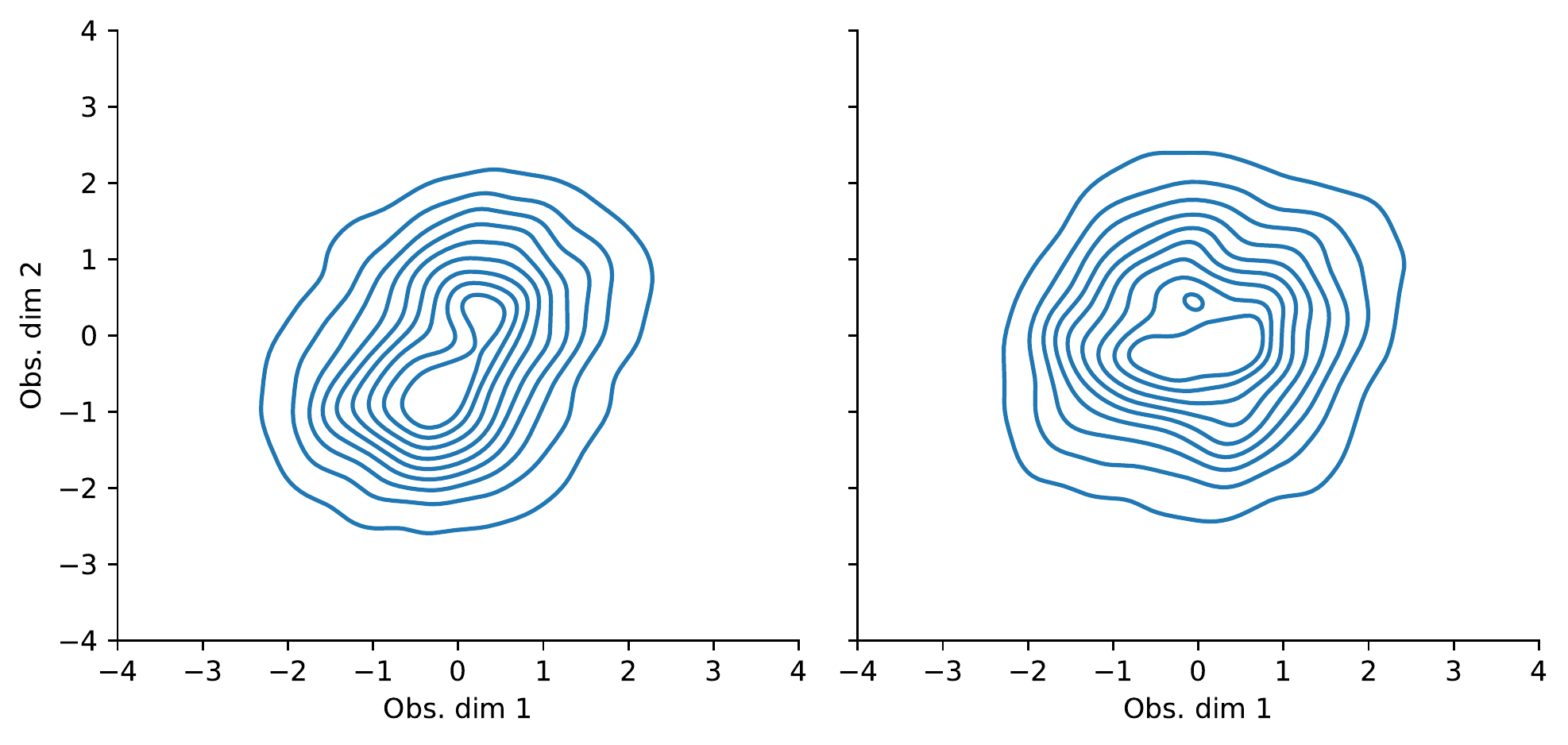}
		\caption{The kernel density estimated distribution of the first two dimensions $R_{t,1:2}$ of the ``oscillating PCA 10'' data. On the left we see the TempVAE model and on the right the historical data. }
		\label{fig:plot_scatter_osc_pca10}
		
	\end{figure}

	For ``oscillating PCA 5'' and ``oscillating PCA 10'', the model identifies two active dimensions through time as can be seen in Figures \ref{fig:plot_Activities_heatmap_osc_pca5} and \ref{fig:plot_Activities_heatmap_osc_pca10}. In our point of view, this is not problematic, since the non-linearities of the artificial neural net can be the reason, that less dimensions are needed to model the data appropriately. This can be observed in Table \ref{tab: NLL oscillating pca}. Comparable scores are achieved for the Score Portfolio NLL. Furthermore, this can also be observed when we look at the fit of the model. For an excerpt of the first two dimensions see Figures \ref{fig:plot_scatter_osc_pca}, \ref{fig:plot_scatter_osc_pca5} and \ref{fig:plot_scatter_osc_pca10}.

	\begin{table}[h!]
		\center
		\begin{tabular}{l|rrr}
			Score & oscPCA2 & oscPCA5 & oscPCA10 \\ \hline
			TempVAE Portfolio NLL & -5.21& -5.77 & -4.80  \\ 
		\end{tabular}
		\caption{The different fit scores for the model TempVAE on the ``oscillating PCA'' data sets. Comparable scores are achieved for the Score Portfolio NLL. The multivariate Scores are not comparable, as the structure of the data is not Gaussian. }
		\label{tab: NLL oscillating pca}
	\end{table}

	\section{Activities on the Stock Market Data and Model Adaptions for High-dimensional Data}\label{app: Activities Stock Market data and High dim model}
	
	\begin{figure}[h!]
		\center
		\includegraphics[scale=0.55]{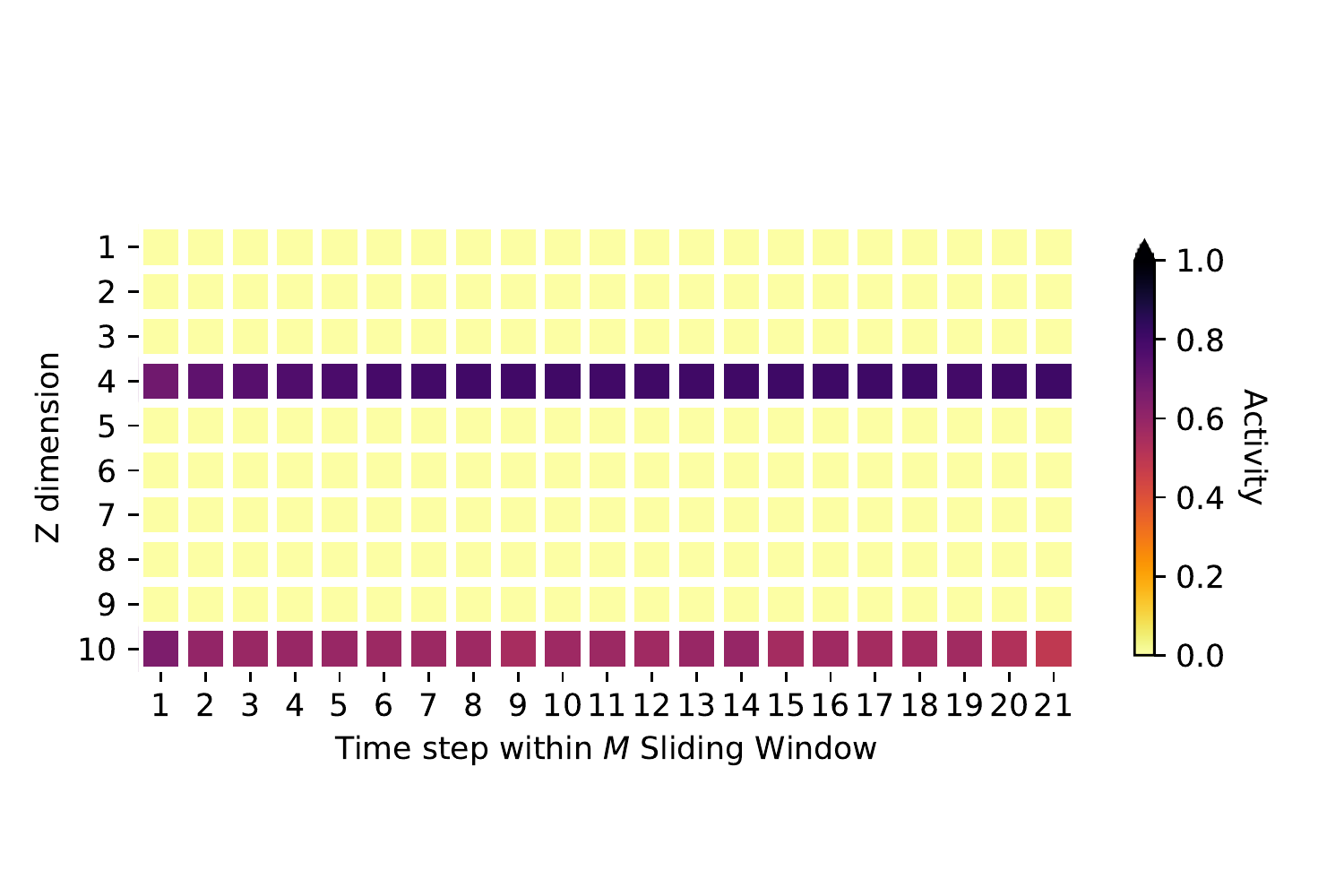}
		\caption{The activity statistics for the ``Oscillating PCA 2'' data. We consider sequences of size $M=21$ as input. Depicted are the activity values of statistic \eqref{eq: time-dep activity} for the $\kappa=10$ dimensional latent space. }
		\label{fig:plot_Activities_heatmap_osc_dax}
		
	\end{figure}
	
\begin{figure}[h!]
		\center
		\includegraphics[scale=0.55]{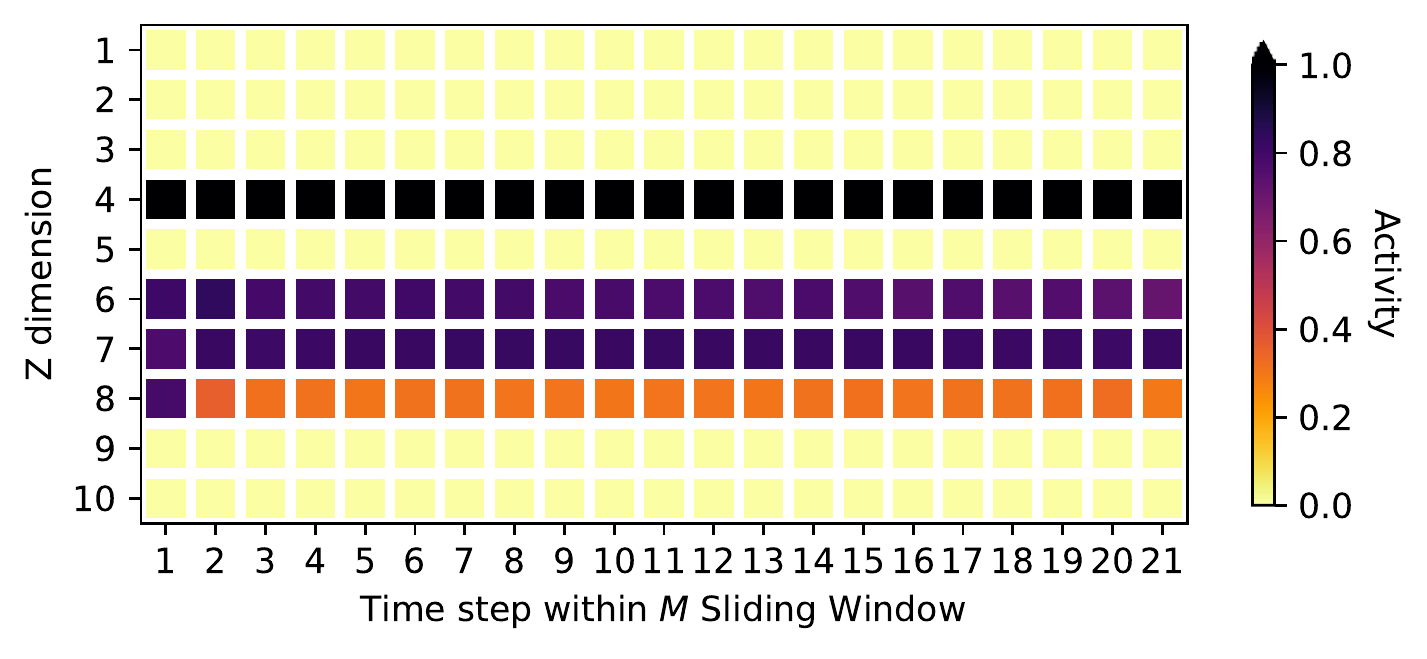}
		\caption{The activity statistics for the ``S\&P500'' data. We consider sequences of size $M=21$ as input. Depicted are the activity values of statistic \eqref{eq: time-dep activity} for the $\kappa=10$ dimensional latent space. }
		\label{fig:plot_Activities_heatmap_SandP}
		
	\end{figure}

	In Figures \ref{fig:plot_Activities_heatmap_osc_dax} and \ref{fig:plot_Activities_heatmap_SandP}, we see the activities for the 22 dimensional ``DAX'' data and the 397 dimensional ``S\&P500'' data.

	As the dimension of the data increased by a factor of 18 compared to the the other data sets in Section \ref{sec:data and preprocessing}, we adjust the model implementation. For the RNN layers in \eqref{eq:generative RNN MLP 2}, \eqref{eq:encoder RNN MLP 2}, \eqref{eq:encoder RNN MLP 3}, \eqref{eq:encoder RNN MLP 4} we choose dimension 80, for the RNN layer in \eqref{eq:prior RNN MLP 2} we choose dimension 16. For the MLPs in \eqref{eq:encoder RNN MLP 1}, \eqref{eq:prior RNN MLP 1} and \eqref{eq:generative RNN MLP 1} we keep the two hidden layers and choose dimensions 60 and 30 for the first and second layer respectively. 
	Furthermore do we reduce the initial learning rate to 1e-4 and change the $\beta$-annealing steps from 20 to 100.

	\section{VaR and Scatterplots}\label{app: VaR plots}
	
	In this section we will show some further VaR path plots for the models TempVAE, HS and DCC-GARCH-MVN. For this see Figures \ref{fig:VaR99 MGARCH,NSVM inset}, \ref{fig:VaR95 MGARCH,NSVM} and \ref{fig:VaR99 MGARCH,NSVM}.
	
\begin{figure}[h!]
		\center
		\includegraphics[scale=0.5]{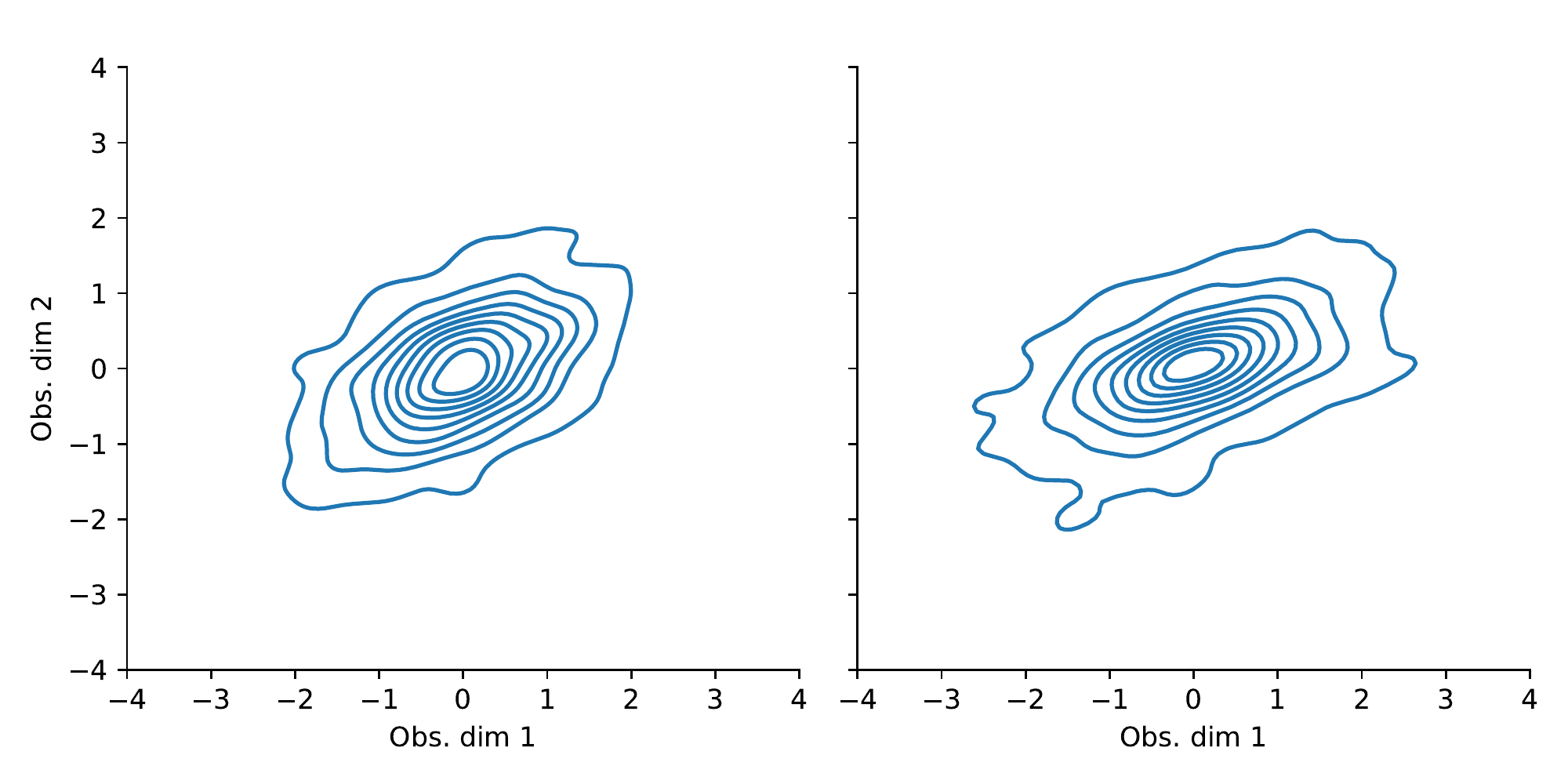}
		\caption{The kernel density estimated distribution of the first two dimensions $R_{t,1:2}$ of the ``DAX'' data. On the left we see the TempVAE model and on the right the historical data. }
		\label{fig:plot_scatter_dax}
		
	\end{figure}

\begin{figure}
		\center
		\includegraphics[scale=0.5]{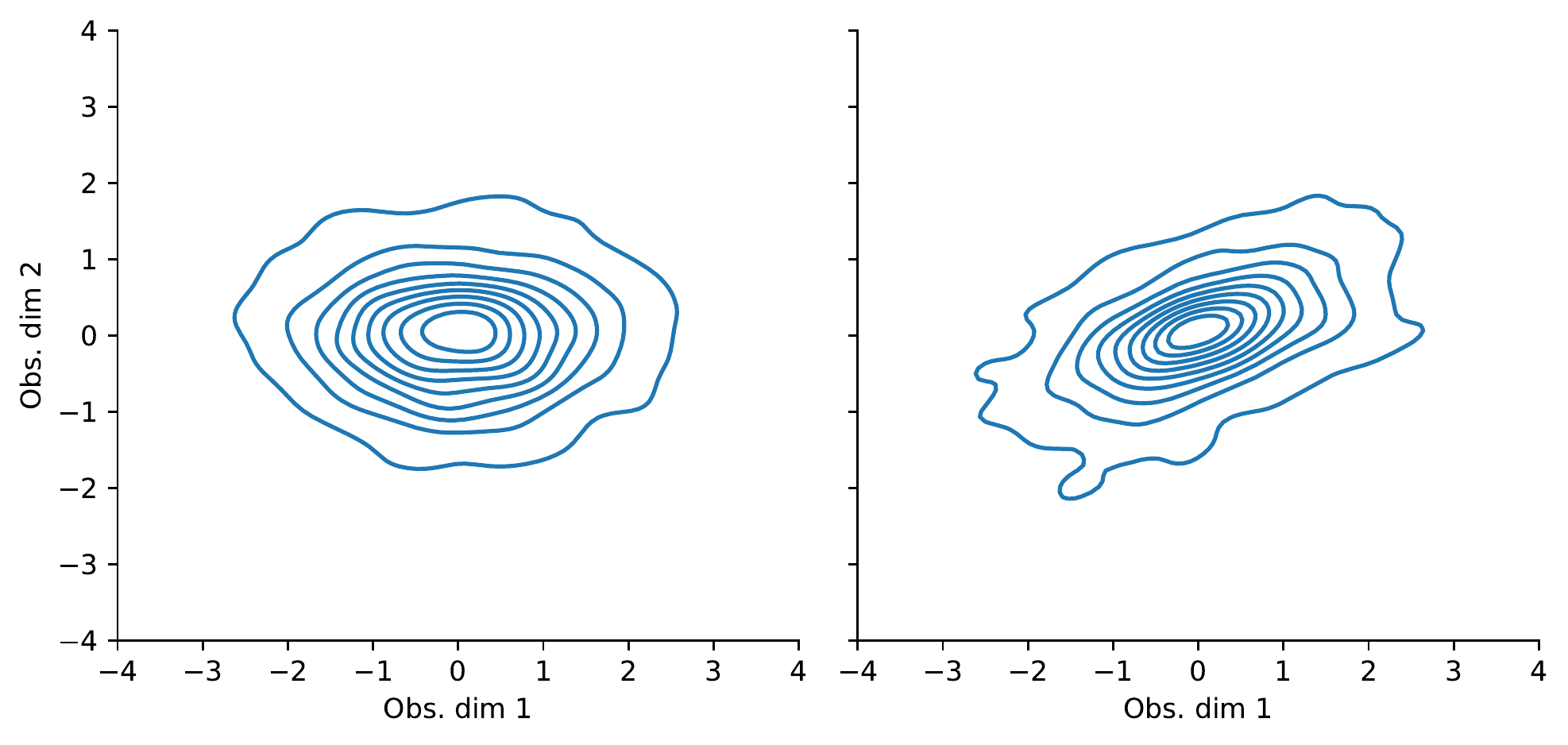}
		\caption{The kernel density estimated distribution of the first two dimensions $R_{t,1:2}$ of the ``DAX'' data. On the left we see the GARCH model and on the right the historical data. }
		\label{fig:plot_scatter_dax GARCH}
		
	\end{figure}

\begin{figure}
		\center
		\includegraphics[scale=0.5]{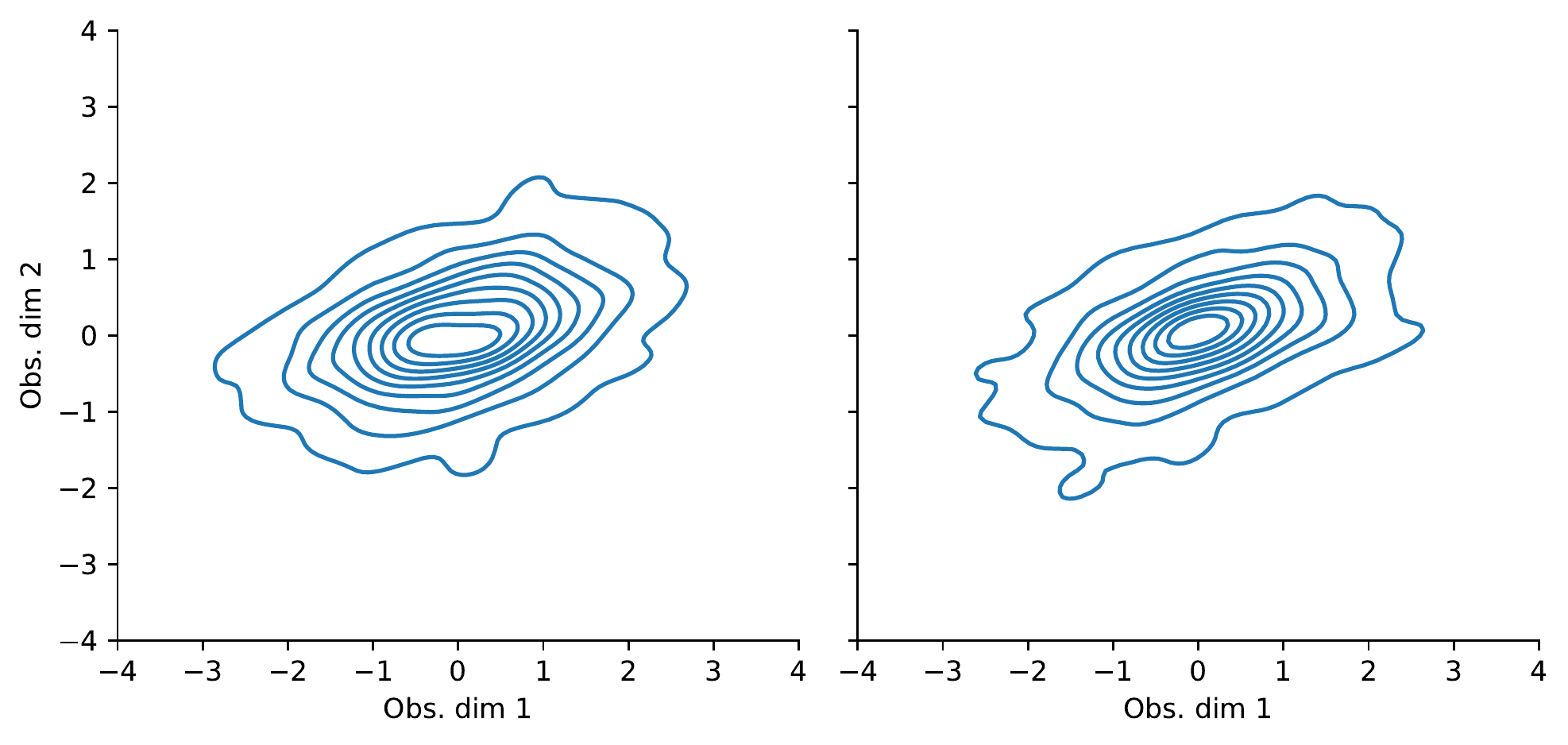}
		\caption{The kernel density estimated distribution of the first two dimensions $R_{t,1:2}$ of the ``DAX'' data. On the left we see the DCC-GARCH-MVN model and on the right the historical data. }
		\label{fig:plot_scatter_dax DCC-GARCH-MVN}
		
	\end{figure}

\begin{figure}
		\center
		\includegraphics[scale=0.5]{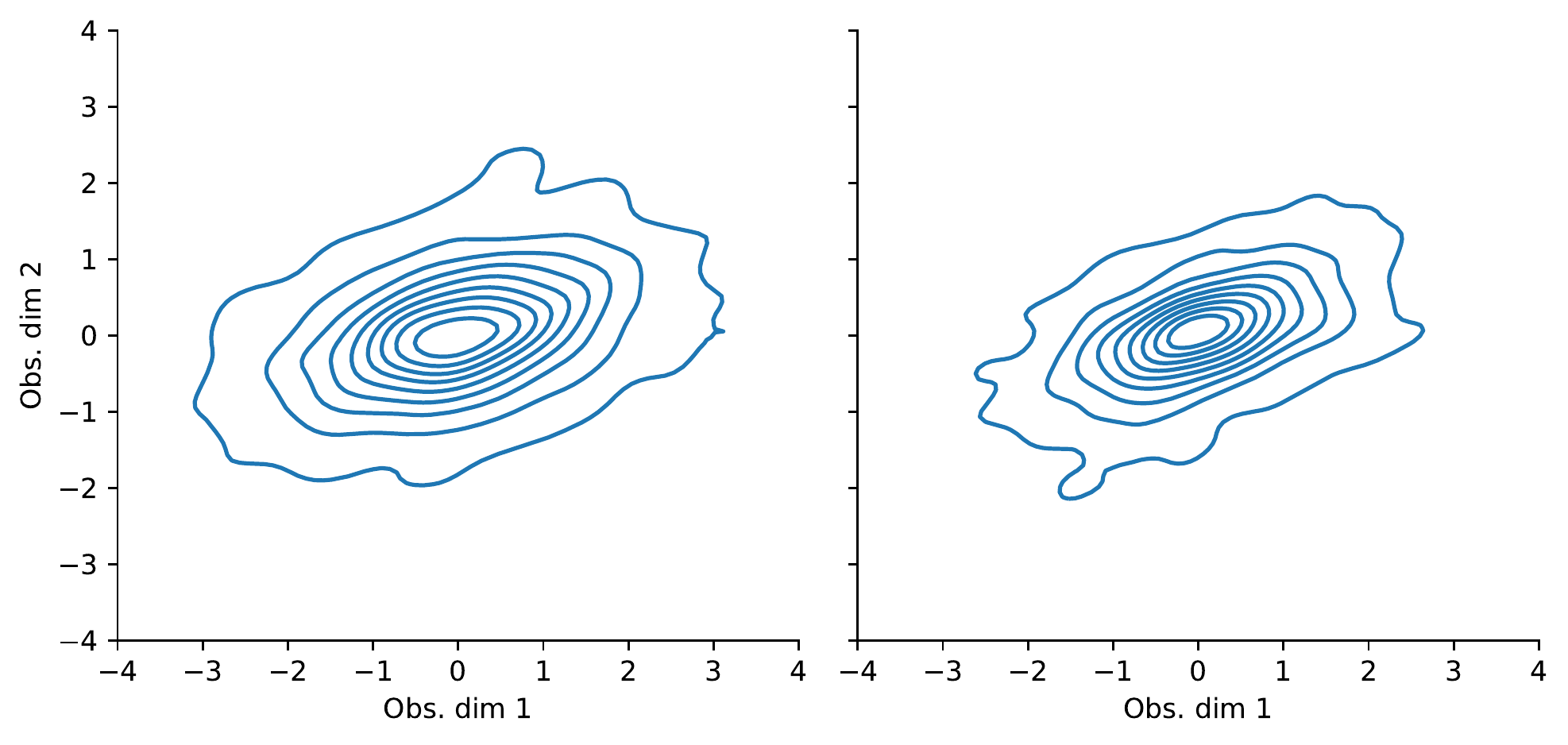}
		\caption{The kernel density estimated distribution of the first two dimensions $R_{t,1:2}$ of the ``DAX'' data. On the left we see the DCC-GARCH-MVt model and on the right the historical data. }
		\label{fig:plot_scatter_dax DCC-GARCH-MVt}
		
	\end{figure}

	Furthermore, we provide Scatterplots for thefirst two dimensions of the different models used for the VaR estimation in the figures \ref{fig:plot_scatter_dax}, \ref{fig:plot_scatter_dax GARCH}, \ref{fig:plot_scatter_dax DCC-GARCH-MVN} and \ref{fig:plot_scatter_dax DCC-GARCH-MVt}.
	\begin{figure}[h!]
		\center
		\includegraphics[scale=0.8]{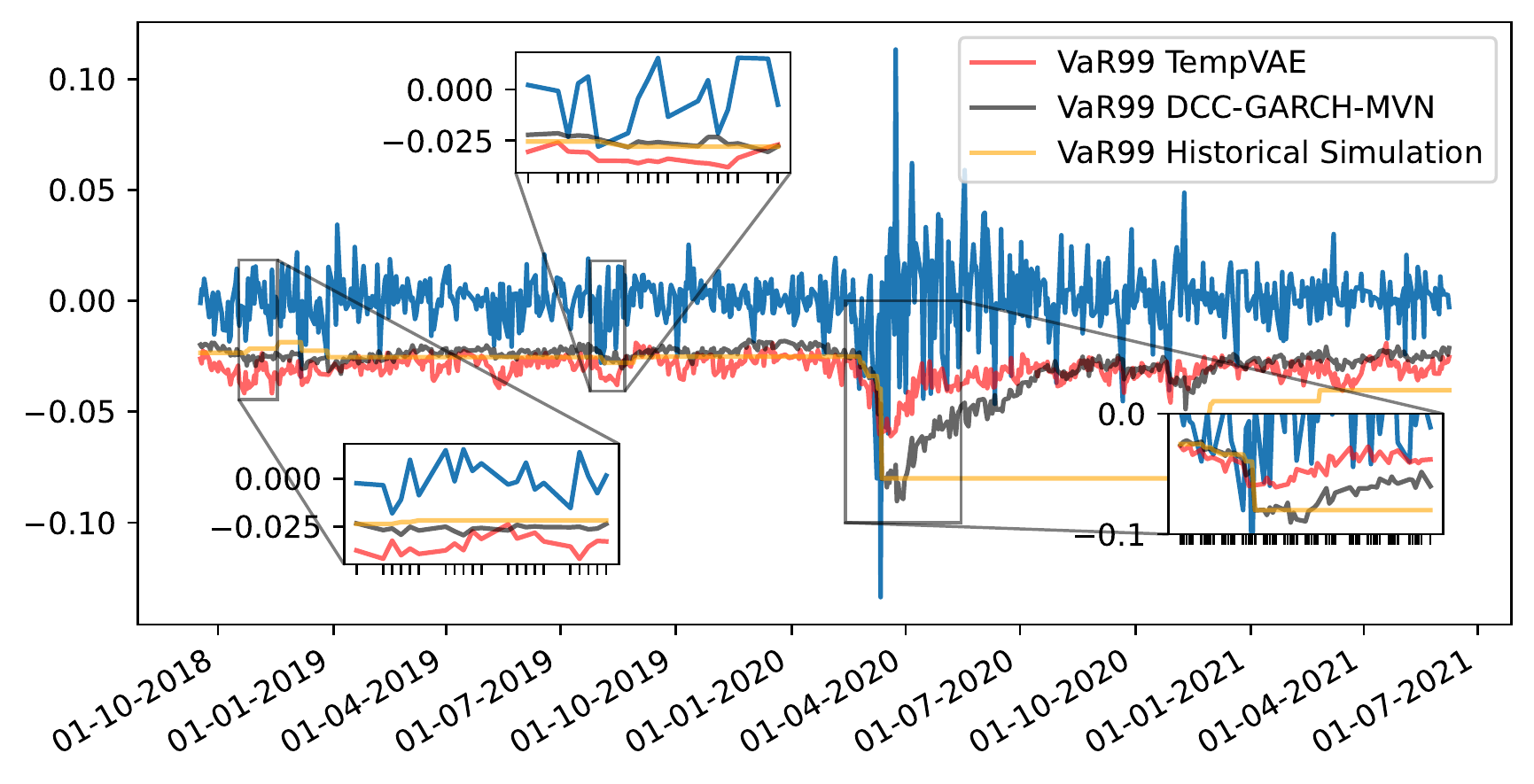}
		\caption{The VaR99 estimates for the models TempVAE, DCC-GARCH-MVN and HS on a fraction of the test data. }
		\label{fig:VaR99 MGARCH,NSVM inset}
		
	\end{figure}
	
	\begin{figure}
		\center
		\includegraphics[scale=0.8]{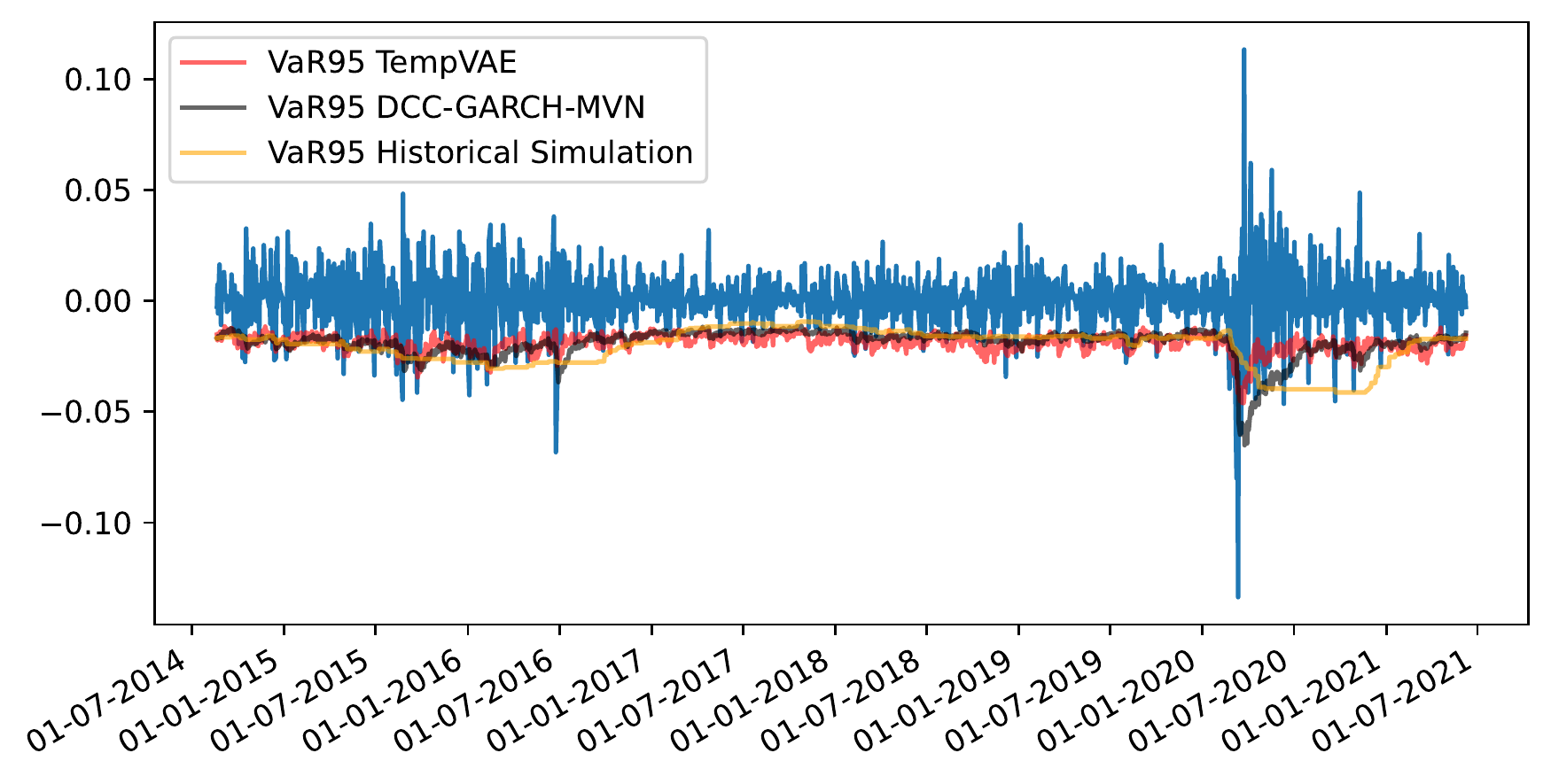}
		\caption{The VaR95 estimates for the two models TempVAE, DCC-GARCH-MVN and HS on the whole test data. }
		\label{fig:VaR95 MGARCH,NSVM}
		
	\end{figure}

	\begin{figure}
		\center
		\includegraphics[scale=0.8]{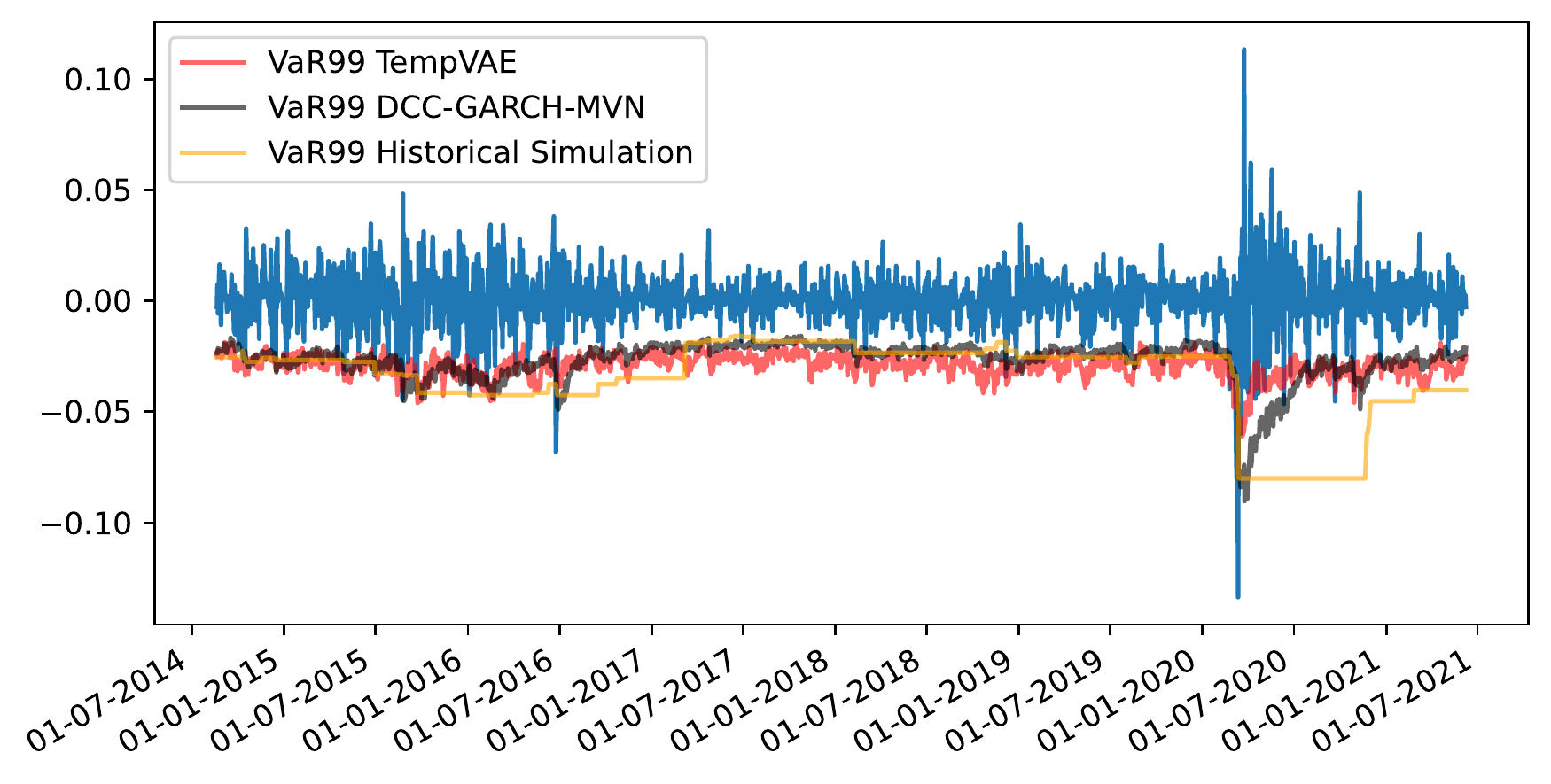}
		\caption{The VaR99 estimates for the two models TempVAE, DCC-GARCH-MVN and HS on the wohle test data. }
		\label{fig:VaR99 MGARCH,NSVM}
		
	\end{figure}
	
\end{document}